\documentclass{article}

\usepackage[final]{neurips_2022}

\usepackage[utf8]{inputenc} 
\usepackage[T1]{fontenc}    
\usepackage{hyperref}       
\usepackage{url}            
\usepackage{booktabs}       
\usepackage{amsfonts}       
\usepackage{nicefrac}       
\usepackage{microtype}      
\usepackage{xcolor}         

\usepackage{graphicx}
\usepackage{amsmath}
\usepackage{multirow}
\usepackage{mathtools}
\usepackage{floatrow}
\usepackage{paracol}
\usepackage{makecell}
\usepackage{csquotes}
\usepackage{comment}

\DeclareMathOperator*{\argmin}{\arg\!\min}

\title{Predictive Coding beyond Gaussian Distributions}

\author{%
    \textbf{Luca Pinchetti}$^{1}$ \ \ \ \  
    \textbf{Tommaso Salvatori}$^{1}$ \ \ \ \  
    \textbf{Yordan Yordanov}$^{1}$ \\
    \textbf{Beren Millidge}$^{2}$ \ \ \ \  
    \textbf{Yuhang Song}$^{1,2,\dag}$ \ \ \ \  
    \textbf{Thomas Lukasiewicz}$^{3,1}$\\
    $^1$\,Department of Computer Science, University of Oxford, UK\\
    $^2$\,MRC Brain Network Dynamics Unit, University of Oxford, UK\\
    $^3$\,Institute of Logic and Computation, TU Wien, Austria  \\
    \texttt{luca.pinchetti@cs.ox.ac.uk, tommaso.salvatori@cs.ox.ac.uk} \\ \texttt{yordan.yordanov@cs.ox.ac.uk, beren.millidge@ndcn.ox.ac.u} \\
    \texttt{yuhang.song@some.ox.ac.uk, thomas.lukasiewicz@cs.ox.ac.uk} \\
}

\begin{document}

\maketitle
\renewcommand{\thefootnote}{\fnsymbol{footnote}}
\footnotetext{ \dag\,Corresponding author.}

\begin{abstract}
  
A large amount of recent research has the far-reaching goal of finding training methods for deep neural networks that can serve as alternatives to backpropagation~(BP). A prominent example is predictive coding (PC), which is a neuroscience-inspired method that performs inference on hierarchical Gaussian generative models. These methods, however, fail to keep up with modern neural networks, as they are unable to replicate the dynamics of complex layers and activation functions. In this work, we solve this problem by generalizing PC to arbitrary probability distributions, enabling the training of architectures, such as transformers, that are hard to approximate with only Gaussian assumptions. We perform three experimental analyses. First, we study the gap between our method and the standard formulation of PC on multiple toy examples. Second, we test the reconstruction quality on variational autoencoders, where our method reaches the same reconstruction quality as BP. Third, we show that our method allows us to train transformer networks and achieve a performance comparable with BP on conditional language models. More broadly, this method allows neuroscience-inspired  learning to be applied to multiple domains, since the internal distributions can be flexibly adapted to the data, tasks, and architectures used.

\end{abstract}

\section{Introduction}

The last decade has seen an explosion of machine learning research fueled by the collection of an unprecedented amount of data and the development of models that can make use of it. Starting with AlexNet \citep{Krizhevsky2012}, deep neural networks trained with \emph{backpropagation (BP)} \citep{rumelhart1986learning} have been established as the best-performing models in many fields \citep{Go_2016, ResNets, GPT3, DALLE, bert}. Despite reaching human-level performance in several tasks \citep{Go_2016,vinyals2017starcraft,vinyals2019grandmaster}, we are still far from artificial general intelligence. The trend has been to constantly increase the number of parameters in such networks, from millions \citep{bert} to hundreds of billions \citep{GPT3}.  The limitations and drawbacks given by the large size of modern architectures have motivated research that looks for alternative methods to train them. 
The direction of research that has inspired this work is that of neuroscience-inspired alternatives to BP, which promise to overcome these drawbacks, due to both interesting properties of their credit assignment, such as plasticity \citep{hebb-organization-of-behavior-1949}, and their biological hardware \citep{kendall2020training}.
These methods have two main advantages relative to standard deep learning models trained with BP. First, it is much more feasible to train them on analog and neuromorphic chips \citep{kendall2020training}. Second, the resulting computational models are extremely flexible in both network layout design and querying techniques \citep{Arbitrary_topology,PC_review,salvatori2021associative,Papadimitriou20}. These two properties could play a crucial role in overcoming the limitations of BP-based learning towards artificial general~intelligence.

\emph{Predictive coding (PC)} is one of the most influential theories of information processing in the brain, initially proposed to explain a large number of brain behaviours \citep{mumford92,friston2009predictive}, and now also a topic of research in machine learning, thanks to the computational model proposed by Rao and Ballard [\citeyear{rao1999pcv}]. This method has in fact been used in supervised and unsupervised tasks \citep{ororbia20,whittington2017approximation,han2018deep}, with an important theoretical result: the original formulation is able to approximate the weight update of BP \citep{whittington2017approximation, millidge2020predictive}, and to exactly replicate it when introducing small variations \citep{ZIL,song2020can}. These results are important, as they draw a strong connection with the aforementioned results obtained by BP in the last decade. PC, however, also presents several interesting properties that make it different from BP: it has an energy-based formulation that allows the design of powerful associative memory models \citep{salvatori2021associative} and to train graphs of any topology \citep{Arbitrary_topology}. 

PC can also be studied from an information theory perspective, as it is a hierarchical generative model \citep{rao1999pcv}. This is a strength of the model, as it has allowed such models to achieve a competitive performance to standard models trained with BP on several generative tasks. These results, however, are all obtained on simple models: sequential architectures with element-wise activations and quadratic energy functions. Deep learning, however, has progressed far from those in recent years, and, therefore, it is necessary to evaluate the performance of PC on more up-to-date and complex architectures to obtain a complete comparison of PC against BP. In this paper, we see that the strict Gaussian assumption is limiting when dealing with more complex architectures such as transformers, preventing PC from reaching the performance obtained by BP. Note that this limitation is not unique to PC, but it is shared among all neuroscience-inspired methods: for example, to our knowledge, none of these frameworks has been successfully used to train language models to date.

In this work, we address this problem by generalizing PC to arbitrary distributions. This allows us to both use Gaussian distributions when allowed, and also to deal with intractable ones by approximating them using a sampling scheme. The resulting framework is coherent with the PC theory, as it enables the definition of layer-wise energy functions that represent prediction errors \citep{rao1999pcv,whittington2017approximation}. In standard PC networks, the error is given by the difference between the expected and actual input of a layer; here, it is defined as the KL-divergence between the expected and actual distributions. We show that these formulations are equivalent when using Gaussian distributions, meaning that our proposed framework is a generalization of standard PC. 
The results of this paper are briefly summarized as follows:
\begin{itemize}
    \item We generalize PC beyond the assumption of a Gaussian generative model. This lets us define prediction errors as ``distances'' between arbitrary distributions at each hierarchical layer of a PC network, and derive a novel free-energy objective to train such networks. 
    \item We empirically show that standard PC is ineffective in training models with complex structure and activation functions, as they cannot be approximated by a Gaussian generative model. Our proposed method, instead, significantly outperforms it, reaching a competitive performance with BP in training both variational autoencoders \citep{VAE} and transformers \citep{attention}. This further bridges the gap in performance between state-of-the-art deep learning methods and neuroscience-inspired learning.
\end{itemize}
The rest of this paper is structured as follows. In Section 2, we introduce the notation used throughout the paper and describe the probabilistic interpretation of PC. In Section 3, we propose how to generalize the definition of PC beyond Gaussian generative models. In Section 4, we evaluate the performance of our proposed method in comparison to standard PC and BP.
Finally, in Sections 5 and 6, we discuss related work and provide a conclusion, respectively.

\section{Predictive Coding: A Probabilistic Interpretation}

\begin{figure}
    \centering
    \includegraphics[width=.9\textwidth]{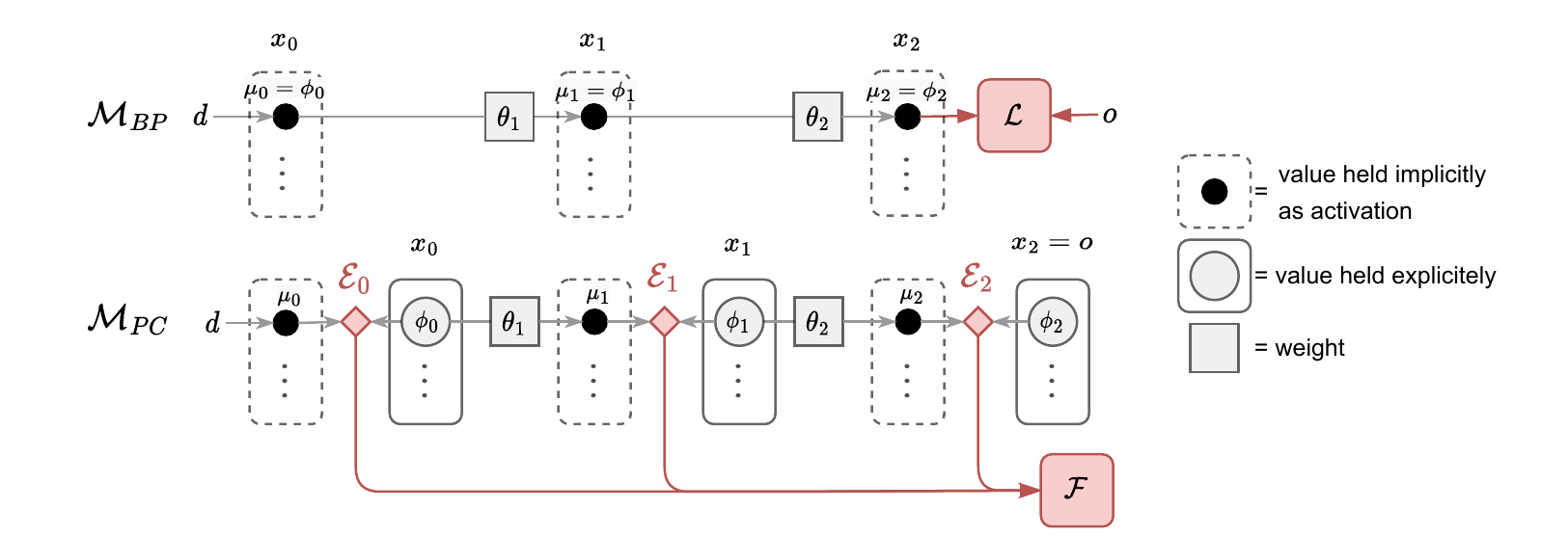}
    \caption{Difference between a network trained with BP (up) and PC (down). The nodes $x_l$ of each layer store the extra parameters $\phi_{l}$. By following the computational arrows backward, the error flows globally from the last layer to the first when using BP. With PC, instead, each layer computes a local error that gets propagated only to nearby nodes.}
    \label{fig:arch_pc_vs_bp}
\end{figure}

In this work, we focus on multi-layer networks organised in a sequence of $L$ layers. These networks define a model $\mathcal{M}_\theta$ that learns the relationship between the input $d$ and the target $o$ by updating its weight parameters $\theta = (\theta_1, \ldots, \theta_L)$. Networks trained with BP have a forward pass, where an output $\hat{o}$ is computed from an input $d$, and a backward pass, where the error given by a loss function $\mathcal{L} = loss(o, \hat{o})$ is propagated back to the network. Every layer of the model has a set of value nodes $x_l$, which contain the value of the prediction $\mu_l$ (i.e., what is normally called activation of a layer), computed accordingly to the value of the previous layer. In detail, we have $\mu_l = f_l(x_{l-1}, \theta_l)$, where $f_l$ is an activation function. In PC, the prediction $\mu_l$ is not stored directly inside the value nodes $x_l$, but rather in separate prediction nodes. The nodes $x_l$ hold, instead, the neural activation  $\phi_l$ of layer $l$. $\phi_l$ is a parameter of the network that can be optimized by minimizing the so-called prediction error between $\mu_l$ and itself. Therefore, a model trained via BP consists exclusively of the parameters $\theta$, while a PC network requires two sets of parameters,  $\theta$ and $ \phi$, where $\phi = (\phi_0, \ldots, \phi_L)$ represents the neural activities of different layers (Fig.~\ref{fig:arch_pc_vs_bp}). Table \ref{tab:notation} summarises the differences between BP and PC.

\begin{table}[h]
    \caption{Clarification of the notation used, highlighting the differences between BP and PC.}
    \label{tab:notation}
    \centering
    \begin{tabular}{c|p{0.3\linewidth}|l|l}
    \toprule
    Method & Value received from $l-1$ & Content of value nodes $x_l$ & Value fed to layer $l+1$\\
    \midrule
    BP & $\mu_l$ & prediction from $l-1$ ($\mu_l$) & $\mu_l$\\
    \midrule
    PC & $\mu_l$ & neural activation of $l$ ($\phi_l$) & $\phi_l$\\
    \bottomrule
    \end{tabular}
\end{table}

\vspace{-1ex}
\subsection{PC as Variational Inference}
PC as a learning algorithm can be mathematically interpreted as a variational inference problem.
It is in fact possible to consider learning as an intractable Bayesian inference process that can be approximated with a tractable optimization problem \citep{FRISTON2003, FRISTON2005, FRISTON2008}. Under this assumption, the neural activities $\phi_i$ in a PC network represent probability distributions.
In particular, Friston based his theory of PC on Gaussian generative models. A detailed review is provided in \citep{PC_review}. 
Assume that we have a generative model $o = g(x)$, where $o$ is a data point and $x$ a set of latent variables, which is described by the joint probability $p(o, x) = p(o|x)p(x)$. We need to solve an inverse problem: given a data point $o$, we need to infer the causes $x$ that generate $o$ through $g$. Similarly to many inverse problems, this one is intractable. In particular, we want to compute the true posterior $p(x|o)$. However, computing it by Bayes rule as $p(x|o) = {p(o,x)}/{p(o)}$ requires the normalizing factor $p(o) = \int p(x,o) dx$, which is, for all but trivial problems, intractable. Variational inference aims to approximate the intractable posterior with a family of distributions $q_\phi(x|o)$, where the parameters $\phi$ have to be learnt. This is generally done via gradient descent on the KL divergence \citep{KL} between the approximated and true posterior. The goal is to compute
\begin{equation}
    q^*_\phi= \argmin_\phi D_{KL}[q_\phi(x|o)||p(x|o)]
\end{equation}
by minimizing an upper bound on the divergence, called the variational free energy $\mathcal{F}$:
\begin{align}
\begin{split}
    \mathcal{F} \coloneqq D_{KL}[q_\phi(x|o)||p(o,x)] \geq D_{KL}[q_\phi(x|o)||p(o,x)] + \ln p(o) = D_{KL}[q_\phi(x|o)||p(x|o)].
\end{split}
\end{align}
The PC framework assumes a Gaussian form for the generative model $p(o,x) = p(o|x)p(x) = \mathcal{N}(o; f(x, \theta), \Sigma_2)\,\mathcal{N}(x, \mu, \Sigma_1)$, where $\Sigma_2$, $\Sigma_1$, and $\mu$ are prior parameters that can optionally be learnt. 
Using as variational posterior the Dirac-delta distribution\footnote{A Gaussian variational posterior under the Laplace approximation can also be used, resulting in the same learning rules as the PC framework proposed here; see \citep{BUCKLEY2017}.}
$q_\phi(x|o) = \delta(x - \phi)$, we get that 
\begin{align}
\begin{split}
    \mathcal{F} &= \mathbb{E}_{q_\phi(x|o)}[\ln q_\phi(x|o)] - \mathbb{E}_{q_\phi(x|o)}[\ln p(o,x)] = - \mathbb{E}_{q_\phi(x|o)}[\ln p(o,x)] = -\ln p(o, \phi),
\end{split}
\end{align}
where the entropy of $q$ is 0. This scheme can be applied to deep neural networks, where $x$ does not represent a homogeneous latent space (e.g., a single layer), but is, instead, organised in a hierarchical structure, defined by the multiple layers of a PC network with widths $w_1, \ldots, w_l$ and nodes $x_0, x_1, \ldots, x_L$. Under this premise, the generative model is as follows:
\begin{align}
\begin{split}
    p(x_{0:L}) & = p(x_0)\prod\nolimits_{l=1}^{L}p(x_l|x_{l-1}) 
    = \mathcal{N}(x_0; \mu_0, \Sigma_0)\prod\nolimits_{l=1}^{L}\mathcal{N}(x_l; \mu_l, \Sigma_l),
    \label{eq:generative_model_gaussians}
\end{split}
\end{align}
where $\mu_l = f_l(x_{l-1}, \theta_l)$, and $x_L$ corresponds to the observation layer and is set to $x_L = o$ during training. The parameters $\Sigma_l$ are prior diagonal covariance matrices, which can be optionally learnt, and $\mu_0$ is an arbitrary prior that can be set to some given data $d$. This is equivalent  to the training of a supervised network with data point $d$ and label $o$.
The energy becomes:
\begin{align}
\begin{split}
    \widetilde{\mathcal{F}} &= -\mathbb{E}_{q_\phi(x_{0:L}|d, o)}[\ln p(x_{0:L})] = \sum\nolimits_{l=0}^L -\ln p(\phi_l|\mu_l)
    = \frac{1}{2}(\sum\nolimits_{l=0}^L \sum\nolimits_{i=1}^{w_l} \Sigma_{l,i}^{-1}\epsilon_{l,i}^2+\ln \Sigma_{l,i}) + k,
    \label{eq:trainablev_pc_energy}
\end{split}
\end{align}
where $k$ is a constant, $\epsilon_l = (\phi_l - \mu_l)$, and $q_\phi(x_{0:L}|d, o) = \prod_{l=0}^L \delta(x_l - \phi_l)$ (implying that $x_l = \phi_l$). In this equation, the total energy is given by the sum of the energies $\mathcal{E}_l$ of every layer, where $\mathcal{E}_l \coloneqq - \ln p(\phi_l|\mu_l)$. 
By assuming identity covariance matrices (i.e., $\Sigma_l = I$)\footnote{Throughout the paper, we assume diagonal covariance matrices for the Gaussian generative model in order to simplify the mathematical derivations. However, our approach can be naturally extended to the case of general covariance matrices \citep{BOGACZ2017}.}, the energy becomes the sum of quadratic errors, introduced by \citet{rao1999pcv}:
\begin{equation}
    \mathcal{F} = \sum\nolimits_{l=0}^L \mathcal{E}_l = \sum\nolimits_{l=0}^L \epsilon_l^2.
    \label{eq:original_pc_energy}
\end{equation}
In most cases, however, the generative model depends on a set of parameters $\theta$: $p(x_0, \ldots, x_L; \theta)$ that need to be learned according to a specific dataset. This can be done via expectation maximization (EM) \citep{EM}, where we first infer the best possible latent variables $\phi$ given a data point $o$ (E-step) and then use them to update the parameters $\theta$ (M-step). In practice, both these phases are  achieved using gradient descent to minimize $\mathcal{F}$.
%
In detail, given a  labelled point $(d, o)$, the input layer prior is set to $\mu_0 = d$, while the output layer nodes are fixed to $\phi_L = o$ for both the inference phase (E-step) and weight update (M-step). During the inference phase, the weight parameters are fixed, and the node values $\phi$ are continuously updated via gradient descent to minimize the energy $\mathcal{F}$. This process either runs until convergence or for a fixed number of steps $T$. When the inference phase ends, a weight update is performed as follows: the node values $\phi$ are fixed, and the weights are updated once via gradient descent on the same energy function $\mathcal{F}$. 


\vspace{-1ex}
\section{Generalization to Arbitrary Distributions}
In this section, we go beyond the strict Gaussian assumption of PC. According to Eq.~\ref{eq:trainablev_pc_energy}, we have that $\mathcal{E}_l = - \ln p(\phi_l|\mu_l) = - \ln \mathcal{N}(\phi_l; f_l(\phi_{l-1}, \theta_{l}), \Sigma_l))$. We can highlight the role of each $\phi_i$, by introducing the ${\cdot}^\mathcal{D}$ and ${\cdot}^\mathcal{S}$ superscripts, which, respectively, indicate that a vector value is interpreted as a distribution (i.e., a vector of sufficient statistics that uniquely identifies a probability distribution, such as the mean $u$ of a Gaussian distribution), or as a single sample. The $\cdot^{\mathcal{S}}$, $\cdot^{\mathcal{D}}$ notation does not, in any case, imply any transformation of the value itself. We get that 
\begin{align}
    \mathcal{E}_l = - \ln \mathcal{N}(\phi_l^\mathcal{S}; f_l(\phi_{l-1}^\mathcal{D}, \theta_{l}), \Sigma_l)) = - \ln p(\phi_l^\mathcal{S}|\phi_{l-1}^\mathcal{D}, \theta_l, \Sigma_l).
\end{align}
Thus, neural activation $\phi_l$ is simultaneously interpreted both as $\phi_l^\mathcal{S}$ and $\phi_l^\mathcal{D}$. This subtle difference has never been highlighted in standard works using hierarchical Gaussian models, since $\phi_l^\mathcal{S}$ corresponds to the maximum likelihood estimation of the Gaussian distribution $\mathcal{N}(\phi_l^\mathcal{D}, \Sigma_l)$, and thus $\phi_l^\mathcal{S} = \phi_l^\mathcal{D}$.
Assuming a Gaussian form for the generative model could, however, become a limiting factor when considering complex architectures. For example, a layer employing a \textit{softmax} activation cannot be easily approximated through a multi-variate Gaussian, since the function introduces a strong dependency between the nodes (i.e., their values have to sum to $1$). 
As we will show in the next sections, using \textit{softmax} activation functions can hinder the ability of a network to learn using the standard definition of PC, and hence a generalized formulation that goes beyond the standard Gaussian assumption is needed.

\begin{figure}[t]
    \centering
    \includegraphics[width=.75\textwidth]{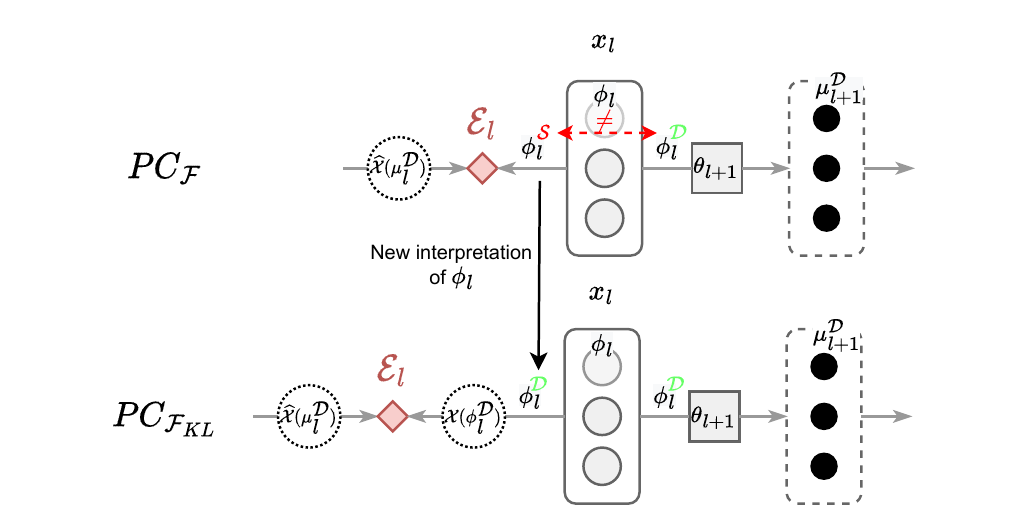}
    \caption{Different layer structure between $PC_{\mathcal{F}}$ (left) and $PC_{\mathcal{F}_{KL}}$ (right). In $PC_{\mathcal{F}}$, the nodes of each layer are simultaneously interpreted as both samples (when evaluating $\mathcal{E}_l$) and as distributions (when computing $\mu_{l+1}$). This inconsistency (highlighted by the dashed red arrow) is not present in $PC_{\mathcal{F}_{KL}}$, where they always represent probability distributions.}
    \label{fig:layer_pc_vs_kl}
\end{figure}

To generalize and extend the canonical formulation of PC, we do the following: instead of directly maximizing the likelihoods $p_l \coloneqq p(\phi_l^\mathcal{S}|\mu_{l}^\mathcal{D})$, we assume that this process is happening repeatedly between each pair of consecutive layers for a single optimization step. We consider the nodes $x_l$ as storing the sufficient statistics $\phi_l^\mathcal{D}$ of an arbitrary distribution $\mathcal{X}_l(\phi_l^\mathcal{D})$, 
\footnote{We use the notation $\widehat{\mathcal{X}}_l(\mu_l)$ to emphasize the dependency of the distribution~$\widehat{\mathcal{X}}_l$ exclusively on the parameters~$\mu_l$. For example, $\widehat{\mathcal{X}}_l$ could be a Gaussian distribution, and $\mu_l$ represents its mean and variance.}
we draw $N$ sample points from it,
$s_l^{(i)} \sim \mathcal{X}_l(\phi_l^\mathcal{D}),\; i \in \{1, \ldots, N\}$,
and we minimize each individual likelihood $p_l^{(i)} = p(s_l^{(i)} | \mu_{l}^\mathcal{D})$.
Furthermore, we remove the Gaussian assumption and consider $\mu_l^\mathcal{D} = f_l(\phi_{l-1}^D, \theta_l)$ to be a parametrization of a generic distribution $\widehat{\mathcal{X}}_l(\mu_l^\mathcal{D})$.
By doing so,
the node values $\phi_l$ are interpreted exclusively as distribution parameters: both the energies $\mathcal{E}_l$ and the activations $\mu_l^\mathcal{D}$ are functions of $\phi_l^\mathcal{D}$. 
It follows that the variational free energy $\mathcal{F}$ is also a function of the expected values given by the likelihoods $p(s_l^{(i)}| \mu_{l}^\mathcal{D}), \, i \in \{1, \ldots, N\}$, for each layer $l$. The energy of each layer is then defined as:
\begin{align}
    \bar{\mathcal{E}}_l \coloneqq -\ln p(\phi_l^\mathcal{D}|\mu_l^\mathcal{D}) \approx \mathcal{H}(\mathcal{X}_l(\phi_l^\mathcal{D}), \widehat{\mathcal{X}}_l(\mu_l^\mathcal{D})).
    \label{eq:fkl_derivation}
\end{align}
A detailed derivation of the above equation is presented in the supplementary material. Knowing that the cross-entropy between two distributions $\mathcal{H}(a,b) = D_{KL}[a || b] + \mathcal{H}(a)$ and that $\mathcal{H}(a) \geq 0$, the total energy of the network can be optimized by minimizing
\begin{equation}
    \mathcal{F}_{KL} = \sum\nolimits_{l=0}^{L} \mathcal{E}_l \coloneqq \sum\nolimits_{l=0}^{L} D_{KL}[\mathcal{X}_l(\phi_l^\mathcal{D}) || \widehat{\mathcal{X}}_l(\mu_l^\mathcal{D})] \leq \sum\nolimits_{l=0}^{L} \mathcal{H}(\mathcal{X}_l(\phi_l^\mathcal{D}), \widehat{\mathcal{X}}_l(\mu_l^\mathcal{D})).
    \label{eq:fkl}
\end{equation}
This follows, as the entropy of $\mathcal{X}_l(\phi_l^\mathcal{D})$ can be assumed as being a constant depending on the training data. Figure \ref{fig:layer_pc_vs_kl} highlights the difference between the original and new PC formulation.
This definition of the variational free energy $\mathcal{F}_{KL}$ does not rely on the Gaussian generative assumption, and (as long as the KL divergence between $\mathcal{X}_l$ and $\widehat{\mathcal{X}}_l$ can be efficiently computed) there is no limit on the kind of distribution that can be used. Throughout our experiments, we assumed that the distributions $\mathcal{X}_l$ and $\widehat{\mathcal{X}}_l$ belong to the same family, but different families can be used in different layers.
In the supplementary material, we show how $\mathcal{F}_{KL}$ is equivalent to $\mathcal{F}$ when assuming a Gaussian generative model. We also analyze the learning dynamics defined by applying the expectation-maximization (EM) algorithm to Eq.~(\ref{eq:fkl}).

%

\vspace{-1ex}
\section{Experiments}
We compare the results of different models trained with BP and PC on various tasks. 
The main goal is to make PC competitive with BP for complex deep neural
architectures. Only recently, PC has begun to be applied to train neural networks for classification tasks \citep{whittington2017approximation}, with similar performance but a greater versatility compared to BP \citep{Arbitrary_topology}. We show that our extended probabilistic formulation can be applied to more complex architectures. 
We refer to each specific version of PC by specifying its energy function.

\begin{figure}[ht]
    \centering
    \includegraphics[width=0.8\textwidth]{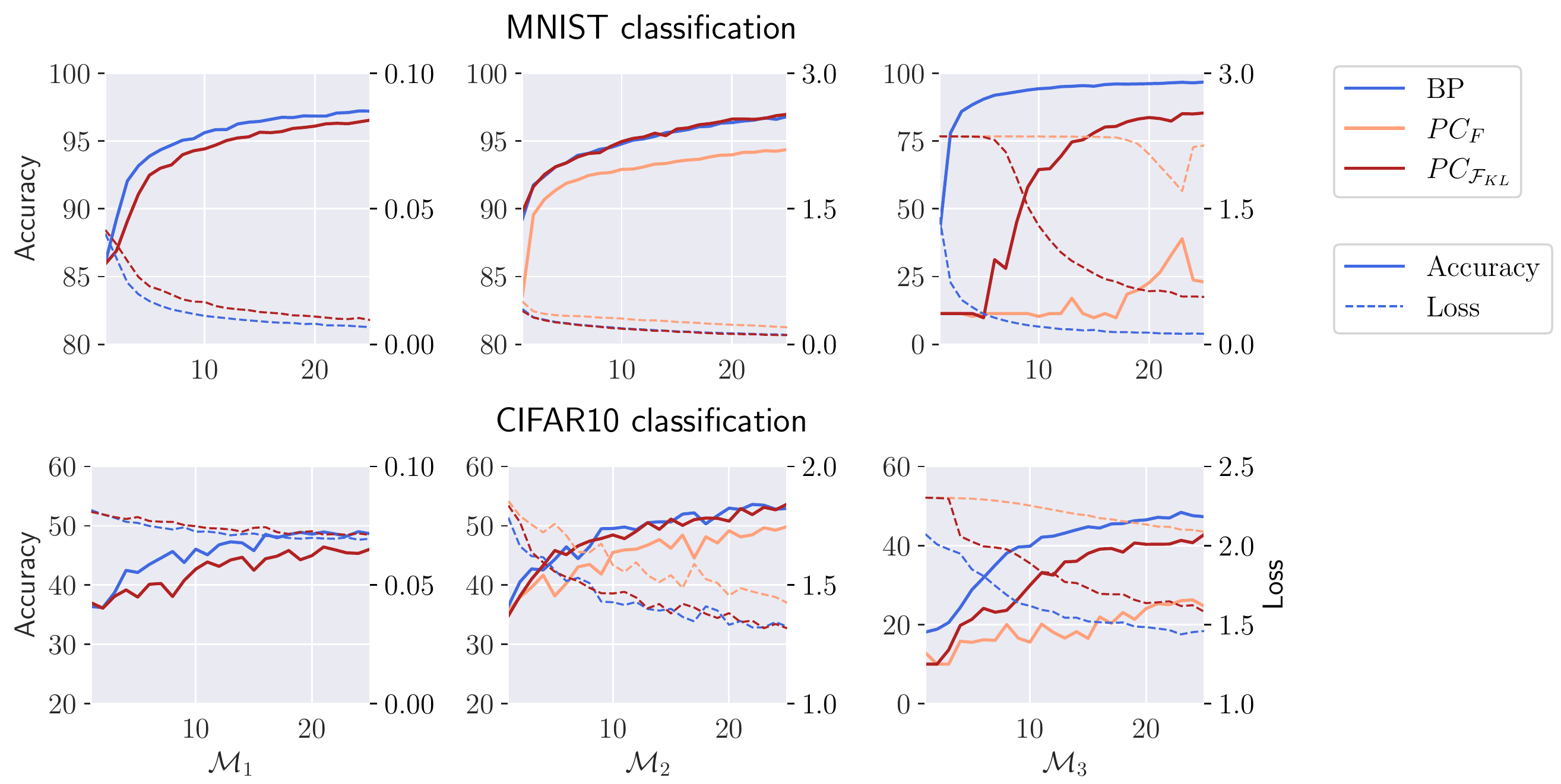}
    \caption{Classification performance of the three models on the MNIST and CIFAR10 datasets. $PC_{\mathcal{F}_{KL}}$ noticeably outperforms $PC_{\mathcal{F}}$, reaching performances comparable with BP. This is true, especially for $\mathcal{M}_2$, which reflects the most commonly used architecture among the three. The x-axis represents the number of epochs.}
    \label{fig:classification}
\end{figure}

\vspace{-1ex}
\subsection{Classification}
As a proof-of-concept experiment, we have trained different models on classification tasks for both the MNIST \citep{MNIST} and CIFAR10 \citep{CIFAR10} datasets. We evaluated the final accuracy of the model when training with BP compared to PC, as well as the training efficiency, measured as improvements over epochs.

\textbf{Setup:} We defined three variations of a fully connected network with $L = 3$ hidden layers and width~$w = 512$:
\begin{itemize}
    \item $\mathcal{M}_1$ uses \textit{tanh} as activation function for the hidden and final layers. The mean squared error (MSE) is the loss function used for BP and to compute the test loss of both PC and BP. 
    \item $\mathcal{M}_2$ uses the \textit{softmax} activation function for the final layer. Consequently, the loss function used is cross-entropy (CE). This architecture corresponds to the one normally used in classification tasks.
    \item $\mathcal{M}_3$ is a copy of $\mathcal{M}_2$ where the activation function of the second hidden layer is replaced with \textit{softmax}. CE is again the loss function used. 
\end{itemize}
Note that the experiments performed with a \textit{softmax} activation in a hidden layer are merely presented with the goal of empirically validating the effectiveness of our theory, as networks of this kind have never been used in practical tasks.  Effectively, $\mathcal{M}_2$ represents the only widely used architecture.
We used a weight learning rate of $\beta_\theta = 0.0001$ for both PC and BP. For PC, we used $T = 32$ $\phi$-steps. We assumed identity covariance matrices for the Gaussian distributions of the generative model. Consequently, the energies  $\mathcal{F}$ and $\mathcal{F}_{KL}$ differ only for the function used for a \textit{softmax}-activated layer. 
For~$\mathcal{F}_{KL}$, that is
\begin{equation}
    \mathcal{E}_{l_{\text{softmax}}} = D_{KL}[\mathcal{X}_l(\phi_l^\mathcal{D}) || \widehat{\mathcal{X}}_l(\mu_l^\mathcal{D})] = \sum\nolimits_{i=1}^{w_l} (\phi_{l,i}) \cdot \ln (\frac{\phi_{l,i}}{\mu_{l,i}}),
    \label{eq:energy_softmax}
\end{equation}
where $\mathcal{X}_l$ and $\widehat{\mathcal{X}}_l$ are discrete distributions. Therefore, in the model $\mathcal{M}_1$, $PC_{\mathcal{F}_{KL}}$ and $PC_{\mathcal{F}}$ are algorithmically equivalent (see the supplementary material). More details about the hyperparameters that we used are given in the supplementary material.

\textbf{Results:} The results are plotted in Fig.~\ref{fig:classification}. They show that neural networks trained with BP and $PC_{\mathcal{F}_{KL}}$ perform similarly in both $\mathcal{M}_1$ and $\mathcal{M}_2$. The original formulation of PC, on the other hand, performs much worse in all the considered experiments. The first column of plots shows that PC and BP perform similarly when trained with fully Gaussian assumptions.
In the experiment on $\mathcal{M}_2$, the standard definition of PC is performing well, even though it is significantly outperformed by BP and $PC_{\mathcal{F}_{KL}}$. When training on $\mathcal{M}_3$, however, the performances are poor. This again does not happen when training using $PC_{\mathcal{F}_{KL}}$, which obtains performances comparable to those of BP. 
Overall, this clearly indicates that $PC_{\mathcal{F}}$ is not suitable to train neural networks that do not rely uniquely on Gaussian generative models. Instead, we experienced a solid improvement when using the $\mathcal{F}_{KL}$ energy function. It is enough to reach the same accuracy-over-epochs ratio achieved by BP when using $\mathcal{M}_2$. Under more uncommon architectural choices, such as $\mathcal{M}_3$, the ratio is slightly worse in favour of BP, but still decisively better than $PC_{\mathcal{F}}$. The difference is particularly noticeable in the first epochs. We believe that it may be due to a not ideal initialization of the weights for the PC network, which is currently using the default initialization designed for BP networks \citep{he2015delving}. Further research in this direction could improve the training performance.

\begin{figure}
    \centering
    \includegraphics[width=.9\textwidth]{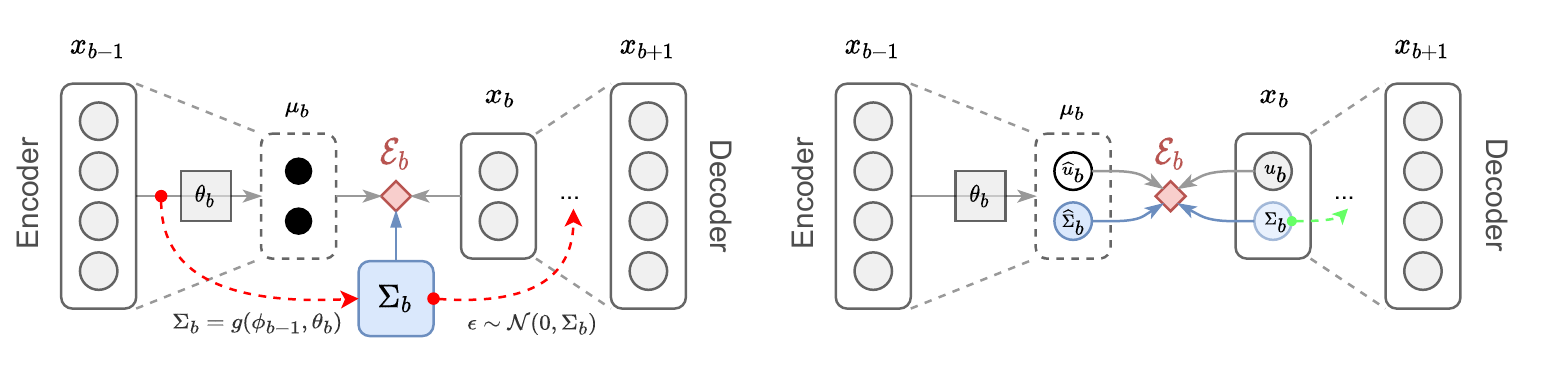}
    \caption{In $PC_{\widetilde{\mathcal{F}}}$ (left), $\Sigma_b$ is stored as an optionally trainable parameter and does not depend on the input $d$. If we were to allow it, and use $\Sigma_b$ to generate $\mu_{b+1}$ (red dashed arrows), we would violate the PC locality assumption, as the error coming from the decoder would flow through $\Sigma_b$ back to the encoder. Using $PC_{\mathcal{F}_{KL}}$ (right), instead, it is possible to have such a dependency by modelling both $\Sigma_b$ and $\widehat{\Sigma}_b$.\vspace*{-1ex}}
    \label{fig:VAE_arch}
\end{figure}

\begin{figure}
    \centering
    \includegraphics[width=.75\textwidth]{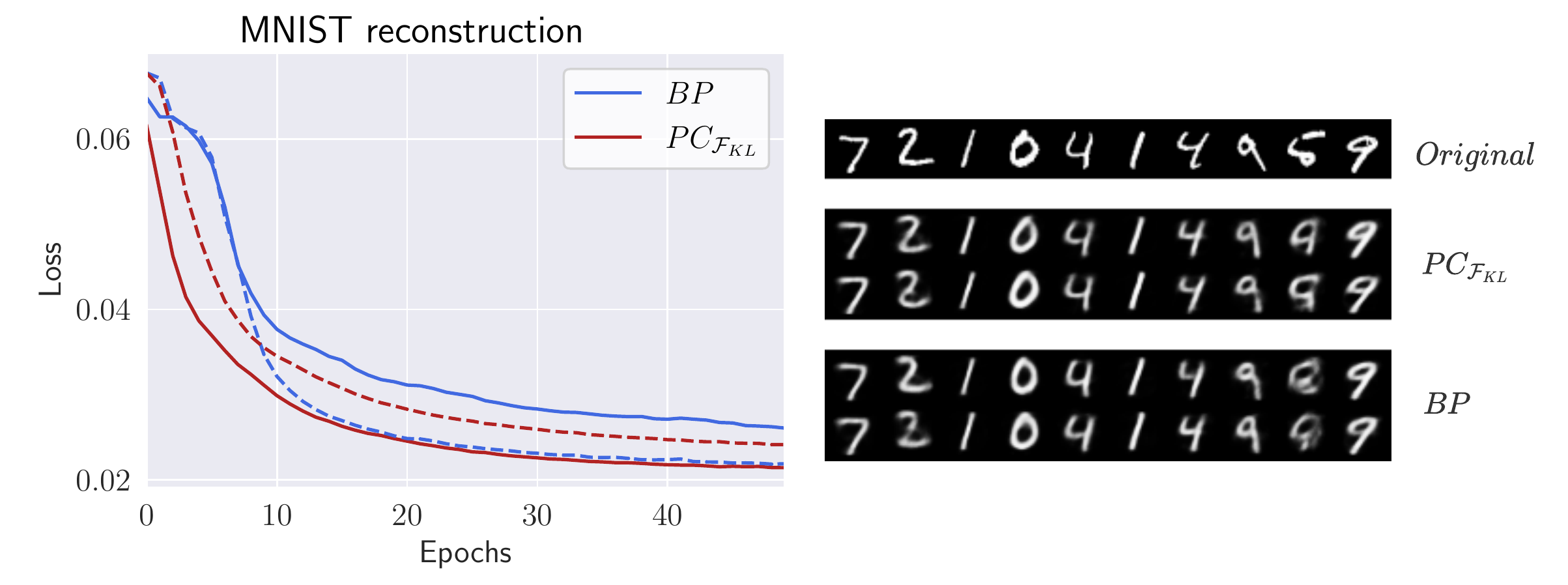}
    \caption{Comparison of BP and PC in training a VAE on the MNIST dataset. The graph on the left shows the test loss over the epochs. The solid and dotted lines represent two different models. Overall PC and BP perform similarly.
    \vspace*{-1ex}}
    \label{fig:vae_mnist}
\end{figure}

\vspace{-1ex}
\subsection{Variational Autoencoders}
Variational autoencoders (VAEs) \citep{VAE} are models that rely on distributions different from Gaussians with fixed covariance matrix. This follows, as the bottleneck layer $b$ of a VAE is required to model the distributions of both the mean and the variance of the latent posterior given a sample $d$, $p(u_b, \Sigma_b | d)$. However, both $PC_{\mathcal{F}}$ and $PC_{\widetilde{\mathcal{F}}}$ are not suitable for that, as they require each layer to represent exclusively the Gaussian mean $u_l$. The optionally learnable parameters $\Sigma_l$ do not depend on the particular input sample $d$. 
Our proposed method, however, overcomes this limitation by learning the full posterior distribution $\mathcal{N}_d(u_b, \Sigma_b)$. This is done by considering the bottleneck layer $b$ as storing the distribution parameters $\phi_b^\mathcal{D} = (u_b, \Sigma_b)$. In this case, $\mu_b^\mathcal{D} = (\widehat{u}_b, \widehat{\Sigma}_b) = f_b(\phi_{b-1}^\mathcal{D}, \theta_{b})$. We then employ the reparametrization trick \citep{VAE} by sampling some Gaussian noise $\bar{\epsilon} \sim \mathcal{N}(0, 1)$ to compute $\mu_{b+1} = f_{b+1}(u_b + \bar{\epsilon} Diag(\Sigma_b^{1/2}), \theta_{b+1})$, which is fed to the next layer. More details are shown in Figure \ref{fig:VAE_arch}.

\textbf{Setup:}
We trained multiple VAEs on the MNIST dataset, comparing the PC and BP training algorithms. Our architectures employed fully connected layers, with \textit{Hardtanh} activation function for the hidden layers and \textit{sigmoid} for the output layer. The bottleneck layer has a total of $w_b = 32\; (= 16 + 16)$ latent units. When training with PC, we assumed Gaussian distributions with identity covariance matrix for all but the bottleneck layer, for which we model both the mean $u_b$ and the diagonal covariance matrix $\Sigma_b$ as explained above. We used $T = 32$, but for each data point we sampled a single $\bar{\epsilon}$ at $t = 0$. We use the same weight learning rate $\beta_\theta = 0.0001$ for both BP and PC. Further details about hyperparameters and implementation details are described in the supplementary material.

\begin{figure}
    \centering
    \includegraphics[width=.87\textwidth]{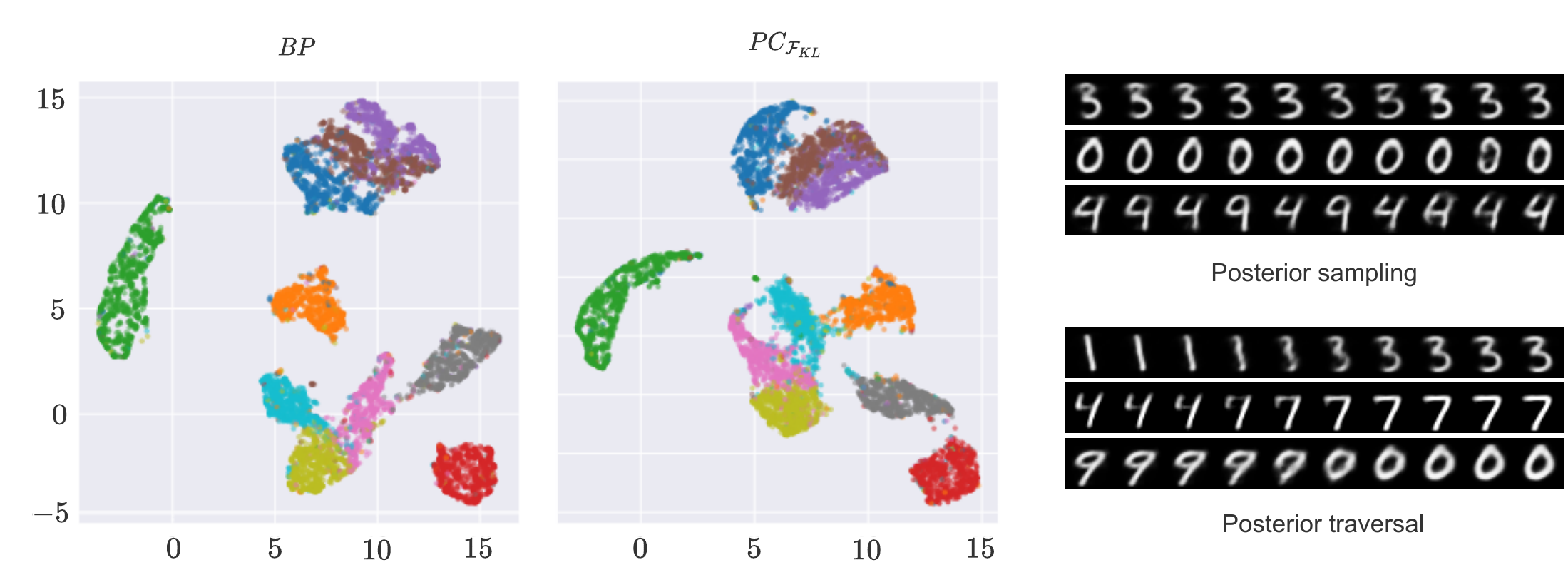}
    \caption{The analysis of the latent space does not highlight any significant differences between PC and BP. UMAP was the algorithm used to obtain the 2D projection \citep{UMAP}. Sampling from the posterior (of the PC-trained model) does not show any anomalies as well.\vspace*{-1ex}}
    \label{fig:vae_latent}
\end{figure}

\textbf{Results:} We observed similar results on a wide range of hyperparameters. In Fig.~\ref{fig:vae_mnist}, we report the performance of both training methods on two different architectures. The final test loss is overall similar, with neither method being decisively better than the other. The learning curves are also comparable, despite PC being generally faster than BP in the first training epochs.
By reconstructing the maximum likelihood estimation of some data points, we can observe how all models produce very similar images. We also performed an analysis of the latent space produced by the encoders and did not detect any significant difference between the two training modes. Figure \ref{fig:vae_latent} reports the results. We sampled the latent posterior distribution by encoding a data point $d$ and decoding multiple data points $d'$ obtained by sampling from $\mathcal{N}_d(\mu_b, \Sigma_b)$. To perform a latent traversal, we encoded two different data points, $d_1$ and $d_2$, and reconstructed the maximum likelihood estimation of the vectors obtained by interpolating the two latent embeddings.
\vspace{-1ex}
\subsection{Transformer Language Models}
To show the performance of our method on more complex tasks, we have trained transformer conditional language models based on the BERT architecture \citep{bert}. The conditional language model objective is enforced by modifying the self-attention mechanism with a triangular mask so that no position can attend to any later position in the sequence.

\textbf{Setup:} We use the 1B Word Benchmark dataset \citep{1b_word_benchmark}, from which we randomly sample 200,000 training and 10,000 dev instances. We choose to restrict the model's input length to 32 input tokens from a vocabulary of 8001 tokens generated by the BPE-based SentencePiece tokenizer \citep{KudoR18}. 
We use one transformer block with one head and a hidden size of 128 throughout the model. For the PC networks, we assume Gaussian distributions with identity covariance matrix for all but the layers that employ a \textit{softmax} activation function (i.e., the attention layers \citep{attention}). In the latter case, we assume a categorical distribution for the generative model. Consequently, the energy function for those layers is the one defined in Eq.~(\ref{eq:energy_softmax}). More implementation details and the hyperparameters are given in the supplementary material.

\begin{figure}
    \centering
    \includegraphics[width=.90\textwidth]{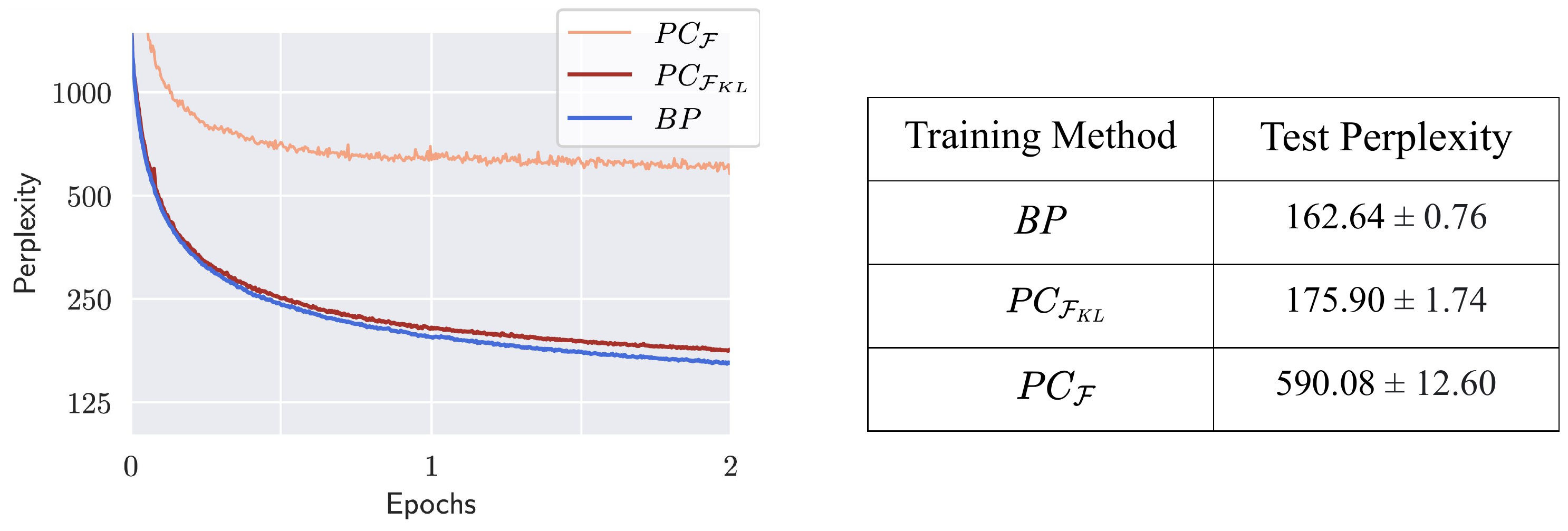}
    \caption{Left: Comparison of language models trained with $BP$ and $PC$, as shown by dev perplexity. Right: Test perplexity achieved by the various training methods for transformer language models. Average ($\pm \sigma$ ) of 10 seeds. \vspace{-1ex}}
    \label{fig:lm_ppl_curves}
\end{figure}

\textbf{Results:} For each model, we compare the three training methods $BP$, $PC_\mathcal{F}$, and $PC_{\mathcal{F}_{KL}}$. We found it beneficial to run multiple weight updates for a single training iteration when using $PC_{\mathcal{F}_{KL}}$, but not for $PC_\mathcal{F}$, where it led to instability. We run a hyperparameter search for each training method, select the best models, compare their training curves, and their test performance, and show qualitative examples of model predictions. Figure~\ref{fig:lm_ppl_curves} shows that $PC_{\mathcal{F}_{KL}}$ significantly outperforms $PC_\mathcal{F}$ in terms of test perplexity
, while having a more stable training curve. We believe that this is because of the \textit{softmax} in both the attention mechanism and the output layer of the transformer model. The performance of $PC_{\mathcal{F}_{KL}}$ is close to that of $BP$
and both training curves look stable. In practical terms, the language models trained by $BP$ and $PC_{\mathcal{F}_{KL}}$ do not differ significantly on the test set. In some cases, the predictions given by $PC_{\mathcal{F}_{KL}}$ are closer to the ground truth, e.g., for \enquote{Yet the bank and its executives are still ready
to support specific Democratic [candidates]}, $PC_{\mathcal{F}_{KL}}$ predicts \enquote{leaders} and \enquote{candidates} as top-2 choices. All models show failure in commonsense reasoning, e.g., for \enquote{I’ve been dreaming about this since I was a [child]} they fail to assign \enquote{child} with $> 1\%$ probability, which shows the limitations of small language models. More examples are given in the supplementary material.

\vspace{-1ex}
\section{Memory Consumption and Computational Complexity} 
\vspace{-1ex}

\textbf{Memory consumption}: PC, in contrast to BP, stores $\mu$ and $\phi$ as different variables, which results in some memory overhead. On the contrary, the number of weights and training parameters used does not change between BP and PC. Therefore, if $M_{BP}$ is the memory consumption of training a model using BP, we have that, in general, $M_{PC} < 2 \cdot M_{BP}$. Actual values depend on the architecture and hyperparameters chosen.\\

\textbf{Computational complexity}: the complexity of a single forward pass in terms of the number of operations is comparable between PC and BP. The same can be said for the backward pass. However, in accordance with the EM algorithm, it is necessary to perform multiple updates on the neurons $x$ before updating the network weights $\theta$. This results in a multiplicative factor that can impact performance compared to BP. Nonetheless, from our experiments, we noticed that even a value as low as $T=4$ or $T=2$, where $T$ is the number of updates of the neurons before performing an update of the parameters, is sufficient given the right hyperparameters. In fact, the experiments on the transformer reached the best perplexity with exactly $T=5$. Furthermore, we can take advantage of the features of PC. One of its major strengths is that each layer computation (both in the forward and backward pass) is local and, therefore, can be executed in parallel, removing one of the main bottlenecks of BP when training deep networks (i.e., the computations induced by each layer have to be executed sequentially). Thus, we expect PC to scale well on large architectures and to bring huge improvements on neuromorphic hardware. Finally, it has already been demonstrated that the speed of energy-based networks can be greatly increased by implementing the relaxation on analog hardware \citep{AnalogForoushani2020, hertz1997nonlinear}, potentially resulting in energy-based networks being faster than BP. Thus, one scientific indication of this work is that the “analog-hardware-friendly” PC can have a reasonable performance on transformers, which opens the door to designing fast hardware-implemented transformers.

\vspace{-1ex}
\section{Related Work} 
\vspace{-1ex}

In the last years, an active research direction that lies at the intersection of machine learning and cognitive science focuses on finding training algorithms for deep neural networks that have a degree of biological plausibility while obtaining good results on machine learning benchmarks. The most popular ones are PC \citep{rao1999pcv,whittington2017approximation}, {equilibrium propagation} \citep{scellier2017equilibrium,scellier2018generalization,scellier2019equivalence}, and {target propagation} \citep{lee2015difference,meulemans2020theoretical,ernoult22}. These methods share multiple similarities, both theoretically and in terms of performance. The first two methods, PC and equilibrium propagation, are able to approximate the weight update of BP when provided with a label that is close in distance to the neural activities of the last layer \citep{whittington2017approximation,scellier2019equivalence}. Target propagation fails to have this property, but instead has been shown to approximate Gauss-Newton optimization \citep{meulemans2020theoretical}. However, PC possesses many unique properties that these methods lack. PC networks can in fact be used to produce efficient associative memory models \citep{salvatori2021associative}, have an update mechanism that produces better learning properties than BP under specific conditions \citep{Song2022.05.17.492325}, and allow training on graphs of any topology \citep{Arbitrary_topology}. Furthermore, they have achieved good results in classification \citep{han2018deep}, generation  \citep{ororbia20}, and reinforcement learning \cite{ororbia2022active,ororbia2022backprop}. For a recent survey on these aspects, see \citep{millidge2022predictive}. To conclude, we are not aware of any neuroscience-inspired learning method before this work that is able to generalize to complex tasks such as language~modeling. 

Progress in this direction is promising, as one of the main limitations of modern architectures is that they are extremely computationally expensive to be trained, with large-scale models sometimes requiring hundreds of GPUs for several weeks \citep{GPT3}. On the other hand, significant breakthroughs on neuromorphic and analog hardware have recently been achieved \citep{strukov2008missing,sebastian2020memory}, which can exploit the aforementioned properties of neuroscience-inspired learning methods, as shown in \citep{kendall2020training}, where the authors simulated the training of a multilayer network on an analog chip in an end-to-end fashion. 

There has been a lot of research done towards bridging the gap in performance between state-of-the-art deep learning methods and neuroscience-inspired learning. Both fields can benefit from each other by drawing inspiration from each other's techniques.
In neuroscience, understanding how the brain learns to associate different areas (e.g., visual and motor cortices) to successfully drive behaviour is of fundamental importance \citep{Petreanu2012, Manita2015, Makino2015LearningET, poort2015learning, Pakan2016, Zmarz2016, attinger2017visuomotor}. However, how to correctly modify synapses to achieve this has puzzled neuroscientists for decades. This is often referred to as the synaptic credit assignment problem \citep{rumelhart1986learning, sutton1998introduction, roelfsema2005attention, bengio2014auto, lee2015difference, roelfsema2018control}, for which the BP algorithm provides an elegant solution.

\vspace{-1ex}
\section{Conclusion}
\vspace{-1ex}
The main motivation behind this work was to make PC competitive with BP for complex deep neural architectures. The tasks in this work are among the most popular and important in the field: image generation and language modelling. In the first case, we trained a variational autoencoder. This model is fully Gaussian, but the bottleneck requires explicitly computable variances. While variations of PC with trainable variances are already defined in the literature \citep{PC_review}, they do not allow dependencies between the variance and the input. Rather, they act as a regulariser within the network. Consequently, they have not been used as a sampling scheme in a specific layer of a PC network. In the second case, we trained a transformer model, intractable before by PC networks, due to the presence of attention (and hence \textit{softmax}), and showed results comparable to those of BP. 
Future work includes applying this method to other complex deep learning architectures, with the far-reaching goal of scaling PC to large-scale machine learning tasks and hence further closing the gap with BP-based learning. 

\section*{Acknowledgments}
This work was supported by the Alan Turing Institute under the EPSRC grant EP/N510129/1, 
by the AXA Research Fund, the EPSRC grant EP/R013667/1, the MRC grant MC\textunderscore UU\textunderscore 00003/1, the BBSRC grant BB/S006338/1, and by the EU TAILOR grant. We also acknowledge the use of the EPSRC-funded Tier 2 facility
JADE (EP/P020275/1) and GPU computing support by Scan Computers International Ltd.
Yuhang Song was supported by the China Scholarship Council under the State Scholarship Fund and by a J.P.~Morgan AI Research Fellowship. 

\bibliographystyle{abbrvnat}
\bibliography{refs}

\begin{thebibliography}{64}
\providecommand{\natexlab}[1]{#1}
\providecommand{\url}[1]{\texttt{#1}}
\expandafter\ifx\csname urlstyle\endcsname\relax
  \providecommand{\doi}[1]{doi: #1}\else
  \providecommand{\doi}{doi: \begingroup \urlstyle{rm}\Url}\fi

\bibitem[Attinger et~al.(2017)Attinger, Wang, and
  Keller]{attinger2017visuomotor}
A.~Attinger, B.~Wang, and G.~B. Keller.
\newblock Visuomotor coupling shapes the functional development of mouse visual
  cortex.
\newblock \emph{Cell}, 169\penalty0 (7):\penalty0 1291--1302, 2017.

\bibitem[Bengio(2014)]{bengio2014auto}
Y.~Bengio.
\newblock How auto-encoders could provide credit assignment in deep networks
  via target propagation.
\newblock \emph{arXiv preprint arXiv:1407.7906}, 2014.

\bibitem[Bogacz(2017)]{BOGACZ2017}
R.~Bogacz.
\newblock A tutorial on the free-energy framework for modelling perception and
  learning.
\newblock \emph{Journal of Mathematical Psychology}, 76:\penalty0 198--211,
  2017.

\bibitem[Brown et~al.(2020)Brown, Mann, Ryder, Subbiah, Kaplan, Dhariwal,
  Neelakantan, Shyam, Sastry, Askell, Agarwal, Herbert-Voss, Krueger, Henighan,
  Child, Ramesh, Ziegler, Wu, Winter, Hesse, Chen, Sigler, Litwin, Gray, Chess,
  Clark, Berner, McCandlish, Radford, Sutskever, and Amodei]{GPT3}
T.~Brown, B.~Mann, N.~Ryder, M.~Subbiah, J.~D. Kaplan, P.~Dhariwal,
  A.~Neelakantan, P.~Shyam, G.~Sastry, A.~Askell, S.~Agarwal, A.~Herbert-Voss,
  G.~Krueger, T.~Henighan, R.~Child, A.~Ramesh, D.~Ziegler, J.~Wu, C.~Winter,
  C.~Hesse, M.~Chen, E.~Sigler, M.~Litwin, S.~Gray, B.~Chess, J.~Clark,
  C.~Berner, S.~McCandlish, A.~Radford, I.~Sutskever, and D.~Amodei.
\newblock Language models are few-shot learners.
\newblock In \emph{Advances in Neural Information Processing Systems},
  volume~33, pages 1877--1901, 2020.

\bibitem[Buckley et~al.(2017)Buckley, Kim, McGregor, and Seth]{BUCKLEY2017}
C.~L. Buckley, C.~S. Kim, S.~McGregor, and A.~K. Seth.
\newblock The free energy principle for action and perception: A mathematical
  review.
\newblock \emph{Journal of Mathematical Psychology}, 81:\penalty0 55--79, 2017.

\bibitem[Chelba et~al.(2013)Chelba, Mikolov, Schuster, Ge, Brants, Koehn, and
  Robinson]{1b_word_benchmark}
C.~Chelba, T.~Mikolov, M.~Schuster, Q.~Ge, T.~Brants, P.~Koehn, and
  T.~Robinson.
\newblock One billion word benchmark for measuring progress in statistical
  language modeling.
\newblock \emph{arXiv:1312.3005}, 2013.

\bibitem[Dempster et~al.(1977)Dempster, Laird, and Rubin]{EM}
A.~P. Dempster, N.~M. Laird, and D.~B. Rubin.
\newblock Maximum likelihood from incomplete data via the {EM} algorithm.
\newblock \emph{Journal of the Royal Statistical Society. Series B
  (Methodological)}, 39\penalty0 (1):\penalty0 1--38, 1977.

\bibitem[Deng(2012)]{MNIST}
L.~Deng.
\newblock The mnist database of handwritten digit images for machine learning
  research.
\newblock \emph{IEEE Signal Processing Magazine}, 29\penalty0 (6):\penalty0
  141--142, 2012.

\bibitem[Devlin et~al.(2018)Devlin, Chang, Lee, and Toutanova]{bert}
J.~Devlin, M.-W. Chang, K.~Lee, and K.~Toutanova.
\newblock {BERT: P}re-training of deep bidirectional transformers for language
  understanding.
\newblock \emph{arXiv:1810.04805}, 2018.

\bibitem[Ernoult et~al.(2022)Ernoult, Normandin, Moudgil, Spinney, Belilovsky,
  Rish, Richards, and Bengio]{ernoult22}
M.~Ernoult, F.~Normandin, A.~Moudgil, S.~Spinney, E.~Belilovsky, I.~Rish, B.~A.
  Richards, and Y.~Bengio.
\newblock Towards scaling difference target propagation by learning backprop
  targets.
\newblock \emph{arXiv:2201.13415}, 2022.

\bibitem[Foroushani et~al.(2020)Foroushani, Assaf, Noshahr, Savaria, and
  Sawan]{AnalogForoushani2020}
A.~N. Foroushani, H.~Assaf, F.~H. Noshahr, Y.~Savaria, and M.~Sawan.
\newblock Analog circuits to accelerate the relaxation process in the
  equilibrium propagation algorithm.
\newblock In \emph{2020 IEEE International Symposium on Circuits and Systems
  (ISCAS)}, 2020.

\bibitem[Friston(2003)]{FRISTON2003}
K.~Friston.
\newblock Learning and inference in the brain.
\newblock \emph{Neural Networks}, 16\penalty0 (9):\penalty0 1325--1352, 2003.

\bibitem[Friston(2005)]{FRISTON2005}
K.~Friston.
\newblock A theory of cortical responses.
\newblock \emph{Philosophical Transactions of the Royal Society B: Biological
  Sciences}, 360\penalty0 (1456):\penalty0 815--836, 2005.

\bibitem[Friston(2008)]{FRISTON2008}
K.~Friston.
\newblock Hierarchical models in the brain.
\newblock \emph{PLOS Computational Biology}, 4\penalty0 (11):\penalty0 1--24,
  11 2008.

\bibitem[Friston and Kiebel(2009)]{friston2009predictive}
K.~Friston and S.~Kiebel.
\newblock Predictive coding under the free-energy principle.
\newblock \emph{Philosophical Transactions of the Royal Society B: Biological
  Sciences}, 364\penalty0 (1521):\penalty0 1211--1221, 2009.

\bibitem[Han et~al.(2018)Han, Wen, Zhang, Fu, Culurciello, and
  Liu]{han2018deep}
K.~Han, H.~Wen, Y.~Zhang, D.~Fu, E.~Culurciello, and Z.~Liu.
\newblock Deep predictive coding network with local recurrent processing for
  object recognition.
\newblock \emph{Advances in Neural Information Processing Systems}, 31, 2018.

\bibitem[He et~al.(2015)He, Zhang, Ren, and Sun]{he2015delving}
K.~He, X.~Zhang, S.~Ren, and J.~Sun.
\newblock Delving deep into rectifiers: Surpassing human-level performance on
  imagenet classification.
\newblock \emph{arXiv:1502.01852}, 2015.

\bibitem[He et~al.(2016)He, Zhang, Ren, and Sun]{ResNets}
K.~He, X.~Zhang, S.~Ren, and J.~Sun.
\newblock Deep residual learning for image recognition.
\newblock In \emph{2016 IEEE Conference on Computer Vision and Pattern
  Recognition}, pages 770--778, 2016.

\bibitem[Hebb(1949)]{hebb-organization-of-behavior-1949}
D.~O. Hebb.
\newblock \emph{The Organization of Behavior: {A} Neuropsychological Theory}.
\newblock Wiley, New York, 1949.

\bibitem[Hertz et~al.(1997)Hertz, Krogh, Lautrup, and
  Lehmann]{hertz1997nonlinear}
J.~Hertz, A.~Krogh, B.~Lautrup, and T.~Lehmann.
\newblock Nonlinear backpropagation: {D}oing backpropagation without
  derivatives of the activation function.
\newblock \emph{IEEE Transactions on Neural Networks}, 8\penalty0 (6):\penalty0
  1321--1327, 1997.

\bibitem[Kendall et~al.(2020)Kendall, Pantone, Manickavasagam, Bengio, and
  Scellier]{kendall2020training}
J.~Kendall, R.~Pantone, K.~Manickavasagam, Y.~Bengio, and B.~Scellier.
\newblock Training end-to-end analog neural networks with equilibrium
  propagation.
\newblock \emph{arXiv:2006.01981}, 2020.

\bibitem[Kingma and Welling(2014)]{VAE}
D.~P. Kingma and M.~Welling.
\newblock Auto-encoding variational {B}ayes.
\newblock In \emph{2nd International Conference on Learning Representations},
  2014.

\bibitem[Krizhevsky et~al.()Krizhevsky, Nair, and Hinton]{CIFAR10}
A.~Krizhevsky, V.~Nair, and G.~Hinton.
\newblock {CIFAR-10 (Canadian Institute for Advanced Research)}.

\bibitem[Krizhevsky et~al.(2012)Krizhevsky, Sutskever, and
  Hinton]{Krizhevsky2012}
A.~Krizhevsky, I.~Sutskever, and G.~E. Hinton.
\newblock {ImageNet} classification with deep convolutional neural networks.
\newblock In \emph{26th Annual Conference on Neural Information Processing
  Systems}, 2012.

\bibitem[Kudo and Richardson(2018)]{KudoR18}
T.~Kudo and J.~Richardson.
\newblock {SentencePiece: A} simple and language independent subword tokenizer
  and detokenizer for neural text processing.
\newblock In \emph{Proceedings of the 2018 Conference on Empirical Methods in
  Natural Language Processing, System Demonstrations}, pages 66--71, 2018.

\bibitem[Kullback and Leibler(1951)]{KL}
S.~Kullback and R.~A. Leibler.
\newblock On information and sufficiency.
\newblock \emph{The Annals of Mathematical Statistics}, 22\penalty0
  (1):\penalty0 79--86, 1951.

\bibitem[Lee et~al.(2015)Lee, Zhang, Fischer, and Bengio]{lee2015difference}
D.-H. Lee, S.~Zhang, A.~Fischer, and Y.~Bengio.
\newblock Difference target propagation.
\newblock In \emph{Joint European Conference on Machine Learning and Knowledge
  Discovery in Databases}, pages 498--515. Springer, 2015.

\bibitem[Makino and Komiyama(2015)]{Makino2015LearningET}
H.~Makino and T.~Komiyama.
\newblock Learning enhances the relative impact of top-down processing in the
  visual cortex.
\newblock \emph{Nature Neuroscience}, 18:\penalty0 1116 -- 1122, 2015.

\bibitem[Manita et~al.(2015)Manita, Suzuki, Homma, Matsumoto, Odagawa, Yamada,
  Ota, Matsubara, Inutsuka, Sato, Ohkura, Yamanaka, Yanagawa, Nakai, Hayashi,
  Larkum, and Murayama]{Manita2015}
S.~Manita, T.~Suzuki, C.~Homma, T.~Matsumoto, M.~Odagawa, K.~Yamada, K.~Ota,
  C.~Matsubara, A.~Inutsuka, M.~Sato, M.~Ohkura, A.~Yamanaka, Y.~Yanagawa,
  J.~Nakai, Y.~Hayashi, M.~Larkum, and M.~Murayama.
\newblock A top-down cortical circuit for accurate sensory perception.
\newblock \emph{Neuron}, 86, 05 2015.

\bibitem[McInnes et~al.(2018)McInnes, Healy, and Melville]{UMAP}
L.~McInnes, J.~Healy, and J.~Melville.
\newblock {UMAP: U}niform manifold approximation and projection for dimension
  reduction, 2018.

\bibitem[Meulemans et~al.(2020)Meulemans, Carzaniga, Suykens, Sacramento, and
  Grewe]{meulemans2020theoretical}
A.~Meulemans, F.~Carzaniga, J.~Suykens, J.~Sacramento, and B.~F. Grewe.
\newblock A theoretical framework for target propagation.
\newblock \emph{Advances in Neural Information Processing Systems},
  33:\penalty0 20024--20036, 2020.

\bibitem[Millidge et~al.(2020)Millidge, Tschantz, and
  Buckley]{millidge2020predictive}
B.~Millidge, A.~Tschantz, and C.~L. Buckley.
\newblock Predictive coding approximates backprop along arbitrary computation
  graphs.
\newblock \emph{arXiv preprint arXiv:2006.04182}, 2020.

\bibitem[Millidge et~al.(2021)Millidge, Seth, and Buckley]{PC_review}
B.~Millidge, A.~K. Seth, and C.~L. Buckley.
\newblock Predictive coding: {A} theoretical and experimental review.
\newblock \emph{CoRR}, abs/2107.12979, 2021.

\bibitem[Millidge et~al.(2022)Millidge, Salvatori, Song, Bogacz, and
  Lukasiewicz]{millidge2022predictive}
B.~Millidge, T.~Salvatori, Y.~Song, R.~Bogacz, and T.~Lukasiewicz.
\newblock Predictive coding: Towards a future of deep learning beyond
  backpropagation?
\newblock \emph{arXiv:2202.09467}, 2022.

\bibitem[Mumford(1992)]{mumford92}
D.~Mumford.
\newblock On the computational architecture of the neocortex.
\newblock \emph{Biological Cybernetics}, 66\penalty0 (3):\penalty0 241--251,
  1992.

\bibitem[Ororbia and Kifer(2020)]{ororbia20}
A.~Ororbia and D.~Kifer.
\newblock The neural coding framework for learning generative models.
\newblock \emph{arXiv:2012.03405}, 2020.

\bibitem[Ororbia and Mali(2022{\natexlab{a}})]{ororbia2022active}
A.~Ororbia and A.~Mali.
\newblock Active predicting coding: Brain-inspired reinforcement learning for
  sparse reward robotic control problems.
\newblock \emph{arXiv preprint arXiv:2209.09174}, 2022{\natexlab{a}}.

\bibitem[Ororbia and Mali(2022{\natexlab{b}})]{ororbia2022backprop}
A.~G. Ororbia and A.~Mali.
\newblock Backprop-free reinforcement learning with active neural generative
  coding.
\newblock In \emph{Proceedings of the AAAI Conference on Artificial
  Intelligence}, volume~36, pages 29--37, 2022{\natexlab{b}}.

\bibitem[Pakan et~al.(2016)Pakan, Lowe, Dylda, Keemink, Currie, Coutts, and
  Rochefort]{Pakan2016}
J.~M. Pakan, S.~C. Lowe, E.~Dylda, S.~W. Keemink, S.~P. Currie, C.~A. Coutts,
  and N.~L. Rochefort.
\newblock Behavioral-state modulation of inhibition is context-dependent and
  cell type specific in mouse visual cortex.
\newblock \emph{eLife}, 5:\penalty0 e14985, 2016.

\bibitem[Papadimitriou et~al.(2020)Papadimitriou, Vempala, Mitropolsky,
  Collins, and Maass]{Papadimitriou20}
C.~H. Papadimitriou, S.~S. Vempala, D.~Mitropolsky, M.~Collins, and W.~Maass.
\newblock Brain computation by assemblies of neurons.
\newblock \emph{Proceedings of the National Academy of Sciences}, 117\penalty0
  (25), 2020.

\bibitem[Petreanu et~al.(2012)Petreanu, Gutnisky, Huber, long Xu, O’Connor,
  Tian, Looger, and Svoboda]{Petreanu2012}
L.~Petreanu, D.~A. Gutnisky, D.~Huber, N.~long Xu, D.~H. O’Connor, L.~Tian,
  L.~Looger, and K.~Svoboda.
\newblock {Activity in motor–sensory projections reveals distributed coding
  in somatosensation}.
\newblock \emph{Nature}, 489\penalty0 (7415):\penalty0 299--303, September
  2012.

\bibitem[Poort et~al.(2015)Poort, Khan, Pachitariu, Nemri, Orsolic, Krupic,
  Bauza, Sahani, Keller, Mrsic-Flogel, and Hofer]{poort2015learning}
J.~Poort, A.~G. Khan, M.~Pachitariu, A.~Nemri, I.~Orsolic, J.~Krupic, M.~Bauza,
  M.~Sahani, G.~B. Keller, T.~D. Mrsic-Flogel, and S.~B. Hofer.
\newblock Learning enhances sensory and multiple non-sensory representations in
  primary visual cortex.
\newblock \emph{Neuron}, 86\penalty0 (6):\penalty0 1478--1490, 2015.

\bibitem[Ramesh et~al.(2021)Ramesh, Pavlov, Goh, Gray, Voss, Radford, Chen, and
  Sutskever]{DALLE}
A.~Ramesh, M.~Pavlov, G.~Goh, S.~Gray, C.~Voss, A.~Radford, M.~Chen, and
  I.~Sutskever.
\newblock Zero-shot text-to-image generation.
\newblock In \emph{Proceedings of the 38th International Conference on Machine
  Learning}, volume 139, pages 8821--8831. PMLR, 2021.

\bibitem[Rao and Ballard(1999)]{rao1999pcv}
R.~Rao and D.~Ballard.
\newblock {Predictive coding in the visual cortex: {A} functional
  interpretation of some extra-classical receptive-field effects}.
\newblock \emph{Nature Neuroscience}, 2:\penalty0 79--87, 1999.

\bibitem[Roelfsema and Holtmaat(2018)]{roelfsema2018control}
P.~R. Roelfsema and A.~Holtmaat.
\newblock Control of synaptic plasticity in deep cortical networks.
\newblock \emph{Nature Reviews Neuroscience}, 19\penalty0 (3):\penalty0 166,
  2018.

\bibitem[Roelfsema and {van Ooyen}(2005)]{roelfsema2005attention}
P.~R. Roelfsema and A.~{van Ooyen}.
\newblock Attention-gated reinforcement learning of internal representations
  for classification.
\newblock \emph{Neural Computation}, 17\penalty0 (10):\penalty0 2176--2214,
  2005.

\bibitem[Rumelhart et~al.(1986)Rumelhart, Hinton, and
  Williams]{rumelhart1986learning}
D.~E. Rumelhart, G.~E. Hinton, and R.~J. Williams.
\newblock Learning representations by back-propagating errors.
\newblock \emph{Nature}, 323\penalty0 (6088):\penalty0 533--536, 1986.

\bibitem[Salvatori et~al.(2021)Salvatori, Song, Hong, Frieder, Sha, Xu, Bogacz,
  and Lukasiewicz]{salvatori2021associative}
T.~Salvatori, Y.~Song, Y.~Hong, S.~Frieder, L.~Sha, Z.~Xu, R.~Bogacz, and
  T.~Lukasiewicz.
\newblock Associative memories via predictive coding.
\newblock In \emph{Advances in Neural Information Processing Systems}, 2021.

\bibitem[Salvatori et~al.(2022{\natexlab{a}})Salvatori, Pinchetti, Millidge,
  Song, Bogacz, and Lukasiewicz]{Arbitrary_topology}
T.~Salvatori, L.~Pinchetti, B.~Millidge, Y.~Song, R.~Bogacz, and
  T.~Lukasiewicz.
\newblock Learning on arbitrary graph topologies via predictive coding.
\newblock \emph{arXiv:2201.13180}, 2022{\natexlab{a}}.

\bibitem[Salvatori et~al.(2022{\natexlab{b}})Salvatori, Song, Xu, Lukasiewicz,
  and Bogacz]{ZIL}
T.~Salvatori, Y.~Song, Z.~Xu, T.~Lukasiewicz, and R.~Bogacz.
\newblock Reverse differentiation via predictive coding.
\newblock In \emph{Proceedings of the 36th AAAI Conference on Artificial
  Intelligence}, 2022{\natexlab{b}}.

\bibitem[Scellier and Bengio(2017)]{scellier2017equilibrium}
B.~Scellier and Y.~Bengio.
\newblock Equilibrium propagation: Bridging the gap between energy-based models
  and backpropagation.
\newblock \emph{Frontiers in Computational Neuroscience}, 11:\penalty0 24,
  2017.

\bibitem[Scellier and Bengio(2019)]{scellier2019equivalence}
B.~Scellier and Y.~Bengio.
\newblock Equivalence of equilibrium propagation and recurrent backpropagation.
\newblock \emph{Neural Computation}, 31\penalty0 (2):\penalty0 312--329, 2019.

\bibitem[Scellier et~al.(2018)Scellier, Goyal, Binas, Mesnard, and
  Bengio]{scellier2018generalization}
B.~Scellier, A.~Goyal, J.~Binas, T.~Mesnard, and Y.~Bengio.
\newblock Generalization of equilibrium propagation to vector field dynamics.
\newblock \emph{arXiv preprint arXiv:1808.04873}, 2018.

\bibitem[Sebastian et~al.(2020)Sebastian, Le~Gallo, Khaddam-Aljameh, and
  Eleftheriou]{sebastian2020memory}
A.~Sebastian, M.~Le~Gallo, R.~Khaddam-Aljameh, and E.~Eleftheriou.
\newblock Memory devices and applications for in-memory computing.
\newblock \emph{Nature Nanotechnology}, 15\penalty0 (7):\penalty0 529--544,
  2020.

\bibitem[Silver et~al.(2016)Silver, Huang, Maddison, Guez, Sifre, van~den
  Driessche, Schrittwieser, Antonoglou, Panneershelvam, Lanctot, Dieleman,
  Grewe, Nham, Kalchbrenner, Sutskever, Lillicrap, Leach, Kavukcuoglu, Graepel,
  and Hassabis]{Go_2016}
D.~Silver, A.~Huang, C.~J. Maddison, A.~Guez, L.~Sifre, G.~van~den Driessche,
  J.~Schrittwieser, I.~Antonoglou, V.~Panneershelvam, M.~Lanctot, S.~Dieleman,
  D.~Grewe, J.~Nham, N.~Kalchbrenner, I.~Sutskever, T.~Lillicrap, M.~Leach,
  K.~Kavukcuoglu, T.~Graepel, and D.~Hassabis.
\newblock Mastering the game of {Go} with deep neural networks and tree search.
\newblock \emph{Nature}, 529\penalty0 (7587):\penalty0 484--489, Jan. 2016.

\bibitem[Song et~al.(2020)Song, Lukasiewicz, Xu, and Bogacz]{song2020can}
Y.~Song, T.~Lukasiewicz, Z.~Xu, and R.~Bogacz.
\newblock Can the brain do backpropagation?—{E}xact implementation of
  backpropagation in predictive coding networks.
\newblock In \emph{Advances in Neural Information Processing Systems},
  volume~33, 2020.

\bibitem[Song et~al.(2022)Song, Millidge, Salvatori, Lukasiewicz, Xu, and
  Bogacz]{Song2022.05.17.492325}
Y.~Song, B.~G. Millidge, T.~Salvatori, T.~Lukasiewicz, Z.~Xu, and R.~Bogacz.
\newblock Inferring neural activity before plasticity: A foundation for
  learning beyond backpropagation.
\newblock \emph{bioRxiv:10.1101/2022.05.17.492325}, 2022.

\bibitem[Strukov et~al.(2008)Strukov, Snider, Stewart, and
  Williams]{strukov2008missing}
D.~B. Strukov, G.~S. Snider, D.~R. Stewart, and R.~S. Williams.
\newblock The missing memristor found.
\newblock \emph{Nature}, 453\penalty0 (7191):\penalty0 80--83, 2008.

\bibitem[Sutton and Barto(1998)]{sutton1998introduction}
R.~S. Sutton and A.~G. Barto.
\newblock \emph{Introduction to Reinforcement Learning}, volume~2.
\newblock MIT Press Cambridge, 1998.

\bibitem[Vaswani et~al.(2017)Vaswani, Shazeer, Parmar, Uszkoreit, Jones, Gomez,
  Kaiser, and Polosukhin]{attention}
A.~Vaswani, N.~Shazeer, N.~Parmar, J.~Uszkoreit, L.~Jones, A.~N. Gomez, L.~u.
  Kaiser, and I.~Polosukhin.
\newblock Attention is all you need.
\newblock In \emph{Advances in Neural Information Processing Systems},
  volume~30, 2017.

\bibitem[Vinyals et~al.(2017)Vinyals, Ewalds, Bartunov, Georgiev, Vezhnevets,
  Yeo, Makhzani, K{\"u}ttler, Agapiou, and Schrittwieser]{vinyals2017starcraft}
O.~Vinyals, T.~Ewalds, S.~Bartunov, P.~Georgiev, A.~S. Vezhnevets, M.~Yeo,
  A.~Makhzani, H.~K{\"u}ttler, J.~Agapiou, and J.~Schrittwieser.
\newblock {StarCraft II: A} new challenge for reinforcement learning.
\newblock \emph{arXiv:1708.04782}, 2017.

\bibitem[Vinyals et~al.(2019)Vinyals, Babuschkin, Czarnecki, Mathieu, Dudzik,
  Chung, Choi, Powell, Ewalds, Georgiev, et~al.]{vinyals2019grandmaster}
O.~Vinyals, I.~Babuschkin, W.~M. Czarnecki, M.~Mathieu, A.~Dudzik, J.~Chung,
  D.~H. Choi, R.~Powell, T.~Ewalds, P.~Georgiev, et~al.
\newblock Grandmaster level in {StarCraft II} using multi-agent reinforcement
  learning.
\newblock \emph{Nature}, 575\penalty0 (7782), 2019.

\bibitem[Whittington and Bogacz(2017)]{whittington2017approximation}
J.~C. Whittington and R.~Bogacz.
\newblock An approximation of the error backpropagation algorithm in a
  predictive coding network with local {H}ebbian synaptic plasticity.
\newblock \emph{Neural Computation}, 29\penalty0 (5):\penalty0 1229--1262,
  2017.

\bibitem[Zmarz and Keller(2016)]{Zmarz2016}
P.~Zmarz and G.~Keller.
\newblock Mismatch receptive fields in mouse visual cortex.
\newblock \emph{Neuron}, 92, 10 2016.

\end{thebibliography}

\newpage

\appendix

\section{$\mathcal{F}_{KL}$ Reduces to $\mathcal{F}$ under Gaussian Assumptions}

In this section, we prove that $PC_{\mathcal{F}_{KL}}$ is a generalized version of PC and, therefore, it retains all its properties and previously achieved results when assuming a Gaussian generative model. Particularly, we show that the newly introduced energy function $\mathcal{F}_{KL}$ reduces to the PC original energy formulation $\mathcal{F}$ and, almost exactly, $\widetilde{\mathcal{F}}$, when reintroducing the Gaussian assumptions for the generative model. Consider the KL divergence formulation and a layer $l$. As per Eq. (\ref{eq:fkl}), $\mathcal{E}_l \coloneqq D_{KL}[\mathcal{X}_l(\phi_l) || \widehat{\mathcal{X}}_l(\mu_l)]$. If we assume that $\mathcal{X}_l$ and $\widehat{\mathcal{X}}_l$ are multivariate Gaussian distributions with means $u_l$, $\widehat{u}_l$ and fixed diagonal covariance matrices $\Sigma_l$, $\widehat{\Sigma}_l$, we have that
\begin{equation}
    \mathcal{E}_l = \frac{1}{2} (\sum_{i=1}^{w_l} \frac{\Sigma_{l,i} + (u_{l,i} - \widehat{u}_{l,i})^2}{\widehat{\Sigma}_{l,i}} + \ln \frac{\widehat{\Sigma}_{l,i}}{\Sigma_{l,i}} - 1).
    \label{eq:kl_to_gaussian}
\end{equation}
By setting $\Sigma_l = \widehat{\Sigma}_l = I$, $u_l = \phi_l$, and $\widehat{u}_l = \mu_l$, then
\begin{equation}
    2\mathcal{E}_l = \sum_{i=1}^{w_l} (u_{l,i} - \widehat{u}_{l,i})^2 =  (\mu_l - \phi_l)^2 = \epsilon_l^2,
    \label{eq:pc_equivalence}
\end{equation}
which equals the energy function for each layer used in Eq. (\ref{eq:original_pc_energy}).
If, instead, we assume that $\widehat{\Sigma}_l$ is a learnable parameter associated with layer $l$ (that is,  $\widehat{\Sigma}_l \in \theta_l$, while keeping $\Sigma_l = I$), we obtain:
\begin{align}
\begin{split}
    2\mathcal{E}_l &= (\sum_{i=1}^{w_l} \frac{\Sigma_{l,i} + (u_{l,i} - \widehat{u}_{l,i})^2}{\widehat{\Sigma}_{l,i}} + \ln \widehat{\Sigma}_{l,i} - 1) \\
    &= (\sum_{i=1}^{w_l} \widehat{\Sigma}_{l,i}^{-1}(\epsilon_{l,i}^2 + 1) +\ln \widehat{\Sigma}_{l,i}) + k' 
    = (\sum_{i=1}^{w_l} \widehat{\Sigma}_{l,i}^{-1}(\epsilon_{l,i}'^{\,2}) +\ln \widehat{\Sigma}_{l,i}) + k',
\end{split}
\end{align}
where $k'$ is a constant. By substituting $(\epsilon_{l,i}^2 + 1) = \epsilon_{l,i}'^{\,2}$, $\mathcal{E}_l$ corresponds to the layer energy function obtained in Eq. (\ref{eq:trainablev_pc_energy}). 

\section{Biological Plausibility}

Biological plausibility is a generic concept in the literature, often used to state that a specific family of models behaves similarly to biological neural networks present in our brains. However, different definitions of biological plausibility exist in the literature, and a model can be considered biologically plausible according to some definitions and not others. In what follows, we refer to the definition introduced in \citep{whittington2017approximation}, mostly restricted to local computations and plasticity. We now discuss how our framework fails to satisfy them at the most general level, and how to address this limitation. Particularly, the learning dynamics determined by the $\mathcal{F}_{KL}$ energy respects the PC assumptions defined by \citet{whittington2017approximation} at the layer level. Particularly:
\begin{itemize}
    \item \textit{Local computation}: the activity of each layer depends only on the activities of its input nodes and their synaptic weights (i.e., $\mu_l = f_l(\phi_{l-1}, \theta_l)$).
    
    \item \textit{Local plasticity}: synaptic plasticity only depends on pre and post-synaptic nodes. In fact, to minimize $\mathcal{F}_{KL}$, we take the derivatives
    \begin{align*}
        \frac{\partial \mathcal{F}_{KL}}{\partial \theta} &= \frac{\partial}{\partial \theta} \sum_{l=0}^{L} D_{KL}[\mathcal{X}_l(\phi_l^\mathcal{D}) || \widehat{\mathcal{X}}_l(\mu_l^\mathcal{D})] = \sum_{l=0}^{L} \frac{\partial}{\partial \theta_l} D_{KL}[\mathcal{X}_l(\phi_l^\mathcal{D}) || \widehat{\mathcal{X}}_l(\mu_l^\mathcal{D})], \\
        \intertext{and, analogously,}
        \frac{\partial \mathcal{F}_{KL}}{\partial \phi} &= \sum_{l=0}^{L-1} \frac{\partial}{\partial \phi_l} (D_{KL}[\mathcal{X}_l(\phi_l^\mathcal{D}) || \widehat{\mathcal{X}}_l(\mu_l^\mathcal{D})] + D_{KL}[X_{l+1}(\phi_{l+1}^\mathcal{D}) || \widehat{\mathcal{X}}_{l+1}(\mu_{l+1}^\mathcal{D})])\\ 
        & \;\;\;\; + \frac{\partial}{\partial \phi_L} D_{KL}[X_L(\phi_L^\mathcal{D}) || \widehat{\mathcal{X}}_L(\mu_L^\mathcal{D})],
    \end{align*}
    where the terms of both summations only depends on $\phi_l$, $\phi_{l+1}$, and $\phi_{l-1}$.
\end{itemize}

However, this does not guarantee that the above two properties are satisfied at the neural level, as the exact neural circuit employed within each layer strictly depends on the distribution families chosen for $\widehat{\mathcal{X}}_l$ and $\mathcal{X}_l$.
Consequently, while the original formulation of PC has some degree of biological plausibility, this may not be true in the general proposed framework. This is because we do not set any limit on the complexity of possible distributions. This could also have repercussions on eventual implementations on analog and neuromorphic hardware.  Hence, an interesting open problem is understanding which classes of probability distributions are biologically plausible, and which allow our framework to be implemented on these emergent technologies. Researchers interested in developing biologically plausible models, could then only restrict their study to specific classes of probability distributions. The same applies to researchers interested in implementing models on analog circuits. 

\section{A More Detailed Analysis of Learning Dynamics of $\mathcal{F}_{KL}$}

In what follows, we explicitly derive the update rules for the two classes of distributions discussed in the main body of the paper: categorical distributions and Gaussian distributions with non-fixed variance.

\textbf{Categorical distributions}: A PC layer following a \textit{softmax} activation function represents a categorical distribution over $w_l$ elements. Each node stores a different probability mass value. We have that, at each time step $t$:
\begin{align}
\begin{split}
    \frac{\partial \phi_{l,j}}{\partial t} = - \frac{\partial \mathcal{F}_{KL}}{\partial \phi_{l,j}} &= -\frac{\partial}{\partial \phi_{l,j}} (\mathcal{E}_{l} + \mathcal{E}_{l+1}) \\
    &= -\frac{\partial}{\partial \phi_{l,j}} (\sum_{i=1}^{w_l} (\phi_{l,i}) \cdot \ln (\frac{\phi_{l,i}}{\mu_{l,i}})  + \mathcal{E}_{l+1}) \\
    &= -\ln \phi_{l,j} + \ln \mu_{l,j} - 1 - \frac{\partial \mathcal{E}_{l+1}}{\partial \phi_{l,j}},
\end{split}
\end{align}
where $\mathcal{E}_{l+1} = 0$ when $l = L$. Furthermore,
\begin{align}
\begin{split}
    \frac{\partial \theta_{l,j,k}}{\partial t} = - \frac{\partial \mathcal{F}_{KL}}{\partial \theta_{l,j,k}} &= -\frac{\partial \mathcal{E}_{l}}{\partial \theta_{l,j,k}} \\
    &= -\frac{\partial}{\partial \theta_{l,j,k}} (\sum_{i=1}^{w_l} (\phi_{l,i}) \cdot \ln (\frac{\phi_{l,i}}{\mu_{l,i}})) \\
    &= \mu_{l,j}^{-1} \phi_{l,j} \; \frac{\partial \mu_{l,j}}{\partial \theta_{l,j,k}} \\
    &= \begin{cases}
            (\phi_{l,j}) (\mu_{l,j}) (1 - \mu_{l,j}), & \text{if }j = k\\
            -(\phi_{l,j}) (\mu_{l,j}) (\mu_{l,k}), & \text{otherwise}
        \end{cases}
    ,
\end{split}
\end{align}
where $\mu_l = f_l(\theta_{l} \, \phi_{l-1})$.

\textbf{Gaussian distributions}: As shown in Section 4.2, we can model a full Gaussian distribution $\mathcal{N}(\widehat{u}_l, \widehat{\Sigma}_l)$, with $(\widehat{u}_l, \widehat{\Sigma}_l) = \mu_l = f_l(\phi_{l-1}, \theta_l)$. In this scenario, the layer $l$ parameterises the distribution $\mathcal{N}(u_l, \Sigma_l)$, and $\phi_l = (u_l, \Sigma_l)$. We are, again, assuming diagonal covariance matrices. The dynamics are as follows:

\begin{align}
\begin{split}
    \frac{\partial u_{l,j}}{\partial t} = - \frac{\partial \mathcal{F}_{KL}}{\partial u_{l,j}} &= -\frac{\partial}{\partial u_{l,j}} (\mathcal{E}_{l} + \mathcal{E}_{l+1}) \\
    &= -\widehat{\Sigma}_{l,j}^{-1} \epsilon_{l,j} - \frac{\partial \mathcal{E}_{l+1}}{\partial u_{l,j}}, \\
\end{split}
\end{align}
\begin{align}
\begin{split}
    \frac{\partial \Sigma_{l,j}}{\partial t} = - \frac{\partial \mathcal{F}_{KL}}{\partial \Sigma_{l,j}} &= -\frac{\partial}{\partial \Sigma_{l,j}} (\mathcal{E}_{l} + \mathcal{E}_{l+1}) \\
    &= \frac{1}{2}(\Sigma_{l,j}^{-1} - \widehat{\Sigma}_{l,j}^{-1}) - \frac{\partial \mathcal{E}_{l+1}}{\partial \Sigma_{l,j}}, \\
\end{split}
\end{align}
and
\begin{align}
\begin{split}
    \frac{\partial \theta_{l,j,k}}{\partial t} = - \frac{\partial \mathcal{F}_{KL}}{\partial \theta_{l,j,k}} &= -\frac{\partial \mathcal{E}_{l}}{\partial \theta_{l,j,k}} \\
    &= \begin{cases}
    \widehat{\Sigma}_{l,j}^{-1} \epsilon_{l,j} - \frac{\partial \widehat{u}_{l,j}}{\partial \theta_{l,j,k}} & \text{if } j \leq w_l / 2 \\
    \frac{1}{2}\widehat{\Sigma}_{l,j'}^{-2}(\epsilon_{l,j'}^2 + \Sigma_{l,j'} - \widehat{\Sigma}_{l,j'}) - \frac{\partial \widehat{\Sigma}_{l,j'}}{\partial \theta_{l,j,k}} & \text{otherwise} 
    \end{cases} \\
    &= \begin{cases}
    \widehat{\Sigma}_{l,j}^{-1} \epsilon_l - \frac{\partial f_{l,j}}{\partial \theta_{l,j,k}} & \text{if } j \leq w_l / 2 \\
    \frac{1}{2}\widehat{\Sigma}_{l,j'}^{-2}(\epsilon_{l,j'}^2 + \Sigma_{l,j'}) - \widehat{\Sigma}_{l,j'}^{-1} - \frac{\partial f_{l,j}}{\partial \theta_{l,j,k}} & \text{otherwise} 
    \end{cases},
\end{split}
\end{align}
where $\epsilon_l = (u_l - \widehat{u}_l)$ and $j' = j - w/2$.

\section{Derivations of the Equations Used in this Work}
In this section, we provide more explicit derivations for several of the equations presented in this work. By doing so, we hope to ease a detailed understanding of the mathematical framework that we defined.

\begin{itemize}

\item 
\textbf{Eq.~(\ref{eq:trainablev_pc_energy})}:
\begin{align*}
    \widetilde{\mathcal{F}} &= -\mathbb{E}_{q_\phi(x_{0:L}|d, o)}[\ln p(x_{0:L})] = \sum_{l=0}^L -\ln p(\phi_l|\mu_l) & \text{// Dirac-delta posterior and Eq.~(\ref{eq:generative_model_gaussians})} \\
    &= -\sum_{l=0}^L \ln \mathcal{N}(\phi_l; \mu_l, \Sigma_l) & \text{// Gaussian generative model} \\
    &= \frac{1}{2}(\sum_{l=0}^L \epsilon_l^T\Sigma_l^{-1}\epsilon_l+\ln 2\pi|\Sigma_l|) & \text{// $\epsilon_l = \phi_l - \mu_l$} \\
    &= \frac{1}{2}(\sum_{l=0}^L \sum_{i=1}^{w_l} \Sigma_{l,i}^{-1}\epsilon_{l,i}^2+\ln \Sigma_{l,i}) + k. & \text{// $\Sigma_l$ is a diagonal matrix} \\
\end{align*}

\item
\textbf{Eq.~(\ref{eq:fkl_derivation})}:
\begin{align*}
    \bar{\mathcal{E}}_l &= -\ln p(\phi_l^\mathcal{D}|\mu_l^\mathcal{D}) &\\
    &\coloneqq - \frac{1}{N} \sum_{i=1}^N \ln p(s_l^{(i)}|\mu_{l}^\mathcal{D}) & \text{// by definition}\\
    &\approx -\mathbb{E}_{s_l \sim \mathcal{X}_l(\phi_l^\mathcal{D})}[\ln p(\widehat{\mathcal{X}}_l(\mu_l^\mathcal{D}) = s_l)] & \text{// assuming large N} \\
    &= - \int_{s_l \in dom(\mathcal{X}_l(\phi_l^\mathcal{D}))} p(\mathcal{X}_l(\phi_l^\mathcal{D}) = s_l) \ln p(\widehat{\mathcal{X}}_l(\mu_l^\mathcal{D}) = s_l)\;ds_l \\
    &= \mathcal{H}(\mathcal{X}_l(\phi_l^\mathcal{D}), \widehat{\mathcal{X}}_l(\mu_l^\mathcal{D})). & \text{// definition of $\mathcal{H}$}
\end{align*}

\item
\textbf{Eq.~(\ref{eq:kl_to_gaussian})}
\begin{align*}
    \mathcal{E}_l &= D_{KL}[\mathcal{X}_l(\phi_l^\mathcal{D}) || \widehat{\mathcal{X}}_l(\mu_l^\mathcal{D})] \\
    &= D_{KL}[\mathcal{N}(u_l, \Sigma_l) || \mathcal{N}(\widehat{u_l}, \widehat{\Sigma_l})] \\
    &= -\int \mathcal{N}(x;\,u_l, \Sigma_l) \ln \mathcal{N}(x;\,\widehat{u_l}, \widehat{\Sigma_l})\,dx + \int \mathcal{N}(x;\,u_l, \Sigma_l) \ln \mathcal{N}(x;\,u_l, \Sigma_l)\,dx \\
    &= \sum_{i=1}^{w_l} \frac{1}{2} \ln (2\pi\widehat{\Sigma}_l^2) + \frac{\Sigma^2 + (u -\widehat{u})^2}{2\widehat{\Sigma}_l^2} - \frac{1}{2}(1 + \ln (2\pi\Sigma_l^2))\;\text{ // diagonal covariance matrices}\\
    &= \frac{1}{2} (\sum_{i=1}^{w_l} \frac{\Sigma_{l,i} + (u_{l,i} - \widehat{u}_{l,i})^2}{\widehat{\Sigma}_{l,i}} + \ln \frac{\widehat{\Sigma}_{l,i}}{\Sigma_{l,i}} - 1).\\
    \label{eq:kl_to_gaussian}
\end{align*}

\end{itemize}

\section{Implementation Details}

In this section, we provide a detailed description of the models and parameters needed to reproduce the results presented in this work. Note that our goal was to compare the performance of different training methods. Hence, we do not aim for state-of-the-art results, but rather a comparable performance across the different training methods for each employed architecture.

\subsection{Classification Networks}

We used fully connected feedforward networks composed by a sequence of $L \in \{3,4,5\}$ fully connected layers of width $w \in \{256, 512, 1024\}$. The weights learning rate was set to $\beta_\theta = 0.0001$. We also experimented with different node learning rates $\beta_\phi \in \{0.01, 0.05, 0.025\}$. We used $T = 32$ $\phi$-steps and initialized the node values at $t = 0$ using a forward pass, as suggested by \citet{song2020can}.
We used the Adam optimizer to optimize the weights of the model, while we used a stochastic gradient descent
optimizer for the nodes $x$.
We did not find any relevant differences in the observed relative performance of the three learning methods among the various combinations of hyperparameters tested. The results reported in Fig.~\ref{fig:classification} were obtained with $w = 512$, $L = 3$, and $\beta_\phi = 0.05$.
We found that using \textit{ReLU} instead of \textit{tanh} as activation function significantly reduces the accuracy achieved by PC (at least with the highly-specific architectures used for this task).

\subsection{Variational Autoencoders}

We used fully connected layers for both the encoders and the decoders. We trained several models with $L \in \{2, 3\}$ layers for both encoder and decoder and width $w \in \{256, 512\}$. We used 32 latent units for the bottleneck layer, divided equally to store mean and variance. The activation function used was \textit{tanh}. Learning rates and optimizers are the same used for classification networks. The variance in the results reported is due to different combinations of the hyperparameters chosen to obtain one or the other architecture. In Fig.~\ref{fig:vae_mnist}, we reported the learning curves for two models. The choice was completely random to highlight the comparable performance of BP and PC on a general architecture.

\subsection{Transformer Language Models}

The 8001-token vocabulary is automatically generated based on a portion of the training data and includes the <sos>, <eos>, and <pad> tokens. The input of the model is restricted to sequences of length up to 34, where to the token of the sentence, we prepend the <sos> token and append the <eos> token. The tokens for each batch are further appended to the same length via the <pad> token.

To optimize the weights of the model, the AdamW optimizer is used with default (0.01) weight decay, and each model is trained for two epochs with a batch size of 8. We use a stochastic gradient descent optimizer for the nodes $x$.

Here are the hyperparameter ranges and best values used for each model:

For $BP$: $\beta_\theta \in \{0.0004, 0.0008, 0.0016, 0.0032, 0.0064\}$. Best value: $0.0016$.

For $PC_{\mathcal{F}}$: $T \in \{4, 5, 6, 7, 8\}$, $\beta_\phi \in \{0.001953125, 0.00390625, 0.0078125, 0.015625, 0.03125\}$, $\beta_\theta \in \{0.0002, 0.0004, 0.0008, 0.0016, 0.0032, 0.0064, 0.0128\}$. Best values: $T=4$, $\beta_\phi = 0.015625$, $\beta_\theta = 0.0064$.

For $PC_{\mathcal{F}_{KL}}$: $T \in \{4, 5, 6, 7, 8\}$, $\beta_\phi \in \{0.25, 0.5, 1.0\}$, $\beta_\theta \in \{0.000025$, $0.00005$, $0.0001$, $0.0002$, $0.0004$, $0.0008, 0.0016\}$. Best values: $T=5$, $\beta_\phi = 0.5$, $\beta_\theta = 0.0008$.

The total training time of the hyperparameter search is approximately 94 hours on one Nvidia Titan RTX GPU.

\subsubsection{Qualitative Results}

Table~\ref{tab:lm_predictions} shows example sentence completions given by $BP$, $PC_{\mathcal{F}}$, and $PC_{\mathcal{F}_{KL}}$ along with the probabilities assigned to each prediction. The sentences were selected subjectively from the test dataset based on how interesting they are and cut right before a subjectively interesting word to be predicted.

\begin{table}[t]
    \caption{Top predictions of each model for completing several sentences. The ground-truth completion is given in [brackets]; the model prediction format is: <word> (<probability \%>).}
    \label{tab:lm_predictions}
    \centering
    \scalebox{0.9}{\begin{tabular}{p{0.45\linewidth}|c l|c l|c l}
    \toprule
        Input sentence & \multicolumn{2}{c}{$BP$} & \multicolumn{2}{c}{$PC_{\mathcal{F}}$} & \multicolumn{2}{c}{$PC_{\mathcal{F}_{KL}}$} \\
        \midrule
        \multirow{2}{*}{\makecell[l]{Yet the bank and its executives are still ready \\ to support specific Democratic [candidates]}}
        & leaders & (7.5) & . & (1.0) & leaders & (12.1) \\
        & Party & (7.2) & , & (1.0) & candidates & (7.3) \\
        & candidates & (3.8) & and & (0.6) & presidential & (4.8) \\

        \midrule
        \multirow{2}{*}{\makecell[l]{GMAC started out offering car [loans]}}
        & and & (2.2) & , & (5.3) & sales & (4.4) \\
        & sales & (1.7) & and & (3.1) & products & (2.3) \\
        & , & (1.7) & in & (2.5) & services & (2.0) \\

        \midrule
        \multirow{2}{*}{\makecell[l]{I've been dreaming about this since I was \\ a [child]}}
        & great & (1.6) & lot & (1.1) & " & (1.9) \\
        & " & (1.5) & good & (1.1) & good & (1.2) \\
        & good & (1.2) & very & (0.9) & year & (1.1) \\
        
        \midrule
        \multirow{2}{*}{\makecell[l]{Here is a breakdown of the seven taxes and \\ fees that have been [collected]}}
        & a & (2.3) & a & (4.7) & a & (4.3) \\
        & to & (2.0) & the & (2.3) & to & (2.6) \\
        & in & (1.9) & in & (1.3) & the & (2.4) \\
        
        \midrule
        \multirow{2}{*}{\makecell[l]{Under the plan, Iceland will reimburse \\ the [money]}}
        & first & (1.9) & best & (1.7) & first & (1.9) \\
        & world & (1.0) & first & (1.6) & same & (1.2) \\
        & same & (0.8) & most & (0.7) & world & (1.2) \\
        
        \midrule
        \multirow{2}{*}{\makecell[l]{Aniston and Pitt were still married when \\ Pitt and Jolie made the 2005 [film]}}
        & , & (2.3) & , & (23.0) & . & (10.2) \\
        & . & (1.9) & and & (5.6) & , & (10.0) \\
        & World & (1.6) & . & (5.1) & and & (3.9) \\
        
    \bottomrule
    \end{tabular}}
\end{table}

\end{document}